\documentclass[letterpaper, 10 pt, conference]{ieeeconf}  

\IEEEoverridecommandlockouts                              

\usepackage{hyperref}
\usepackage{graphicx}
\usepackage{graphics} 
\usepackage{mathptmx}      
\usepackage{latexsym}
\usepackage{algorithm}
\usepackage{algorithmicx}
\usepackage{algpseudocode}
\usepackage{subcaption}
\usepackage{amsmath}
\usepackage{epsfig} 
\usepackage{mathptmx} 
\usepackage{times} 
\usepackage{caption}

\title{\LARGE \bf
BTO-RRT: A rapid, optimal, smooth and point cloud-based path planning algorithm
}

\author{Zhaoliang Zheng,~\IEEEmembership{Member,~IEEE,}
        Thomas R. Bewley,
        Falko Kuester,
        and Jiaqi Ma,~\IEEEmembership{Member,~IEEE}
\thanks{Zhaoliang Zheng and Jiaqi Ma are with the Department
of Electrical and Computer Engineering and Civil and Environmental Engineering, University of California, Los Angeles,
CA, 90024 USA e-mail: zhz03@g.ucla.edu.}
\thanks{Thomas R. Bewley and Falko Kuester are with the University of California, San Diego.}
}

\begin{document}

\maketitle
\thispagestyle{empty}
\pagestyle{empty}

\begin{abstract}

This paper explores a rapid, optimal smooth path-planning algorithm for robots (e.g., autonomous vehicles) in point cloud environments. Derivative maps such as dense point clouds, mesh maps, Octomaps, etc. are frequently used for path planning purposes. A bi-directional target-oriented point planning algorithm, directly using point clouds to compute the optimized and dynamically feasible trajectories, is presented in this paper. This approach searches for obstacle-free, low computational cost, smooth, and dynamically feasible paths by analyzing a point cloud of the target environment, using a modified bi-directional and RRT-connect-based path planning algorithm, with a k-d tree-based obstacle avoidance strategy and three-step optimization. This presented approach bypasses the common 3D map discretization, directly leveraging point cloud data and it can be separated into two parts: modified RRT-based algorithm core and the three-step optimization. Simulations on 8 2D maps with different configurations and characteristics are presented to show the efficiency and 2D performance of the proposed algorithm. Benchmark comparison and evaluation with other RRT-based algorithms like RRT, B-RRT, and RRT star are also shown in the paper. Finally, the proposed algorithm successfully achieved different levels of mission goals on three 3D point cloud maps with different densities. The whole simulation proves that not only can our algorithm achieves a better performance on 2D maps compared with other algorithms, but also it can handle different tasks(ground vehicles and UAV applications) on different 3D point cloud maps, which shows the high performance and robustness of the proposed algorithm. The algorithm is open-sourced at \url{https://github.com/zhz03/BTO-RRT}

\end{abstract}

\section{Introduction}
Nowadays, robots and autonomous driving cars play a more and more important role in many areas, such as intelligent transportation systems~\cite{xu2022v2x, xu2022opv2v}. Mobile robots and robotic manipulators are widely used in unknown environment exploration, navigation, localization and mapping, etc\cite{cao2022beaglerover}. To have a better and smarter strategy for performing these tasks, path planning is a necessary and key technique. Path planning algorithms for robots can be defined as finding an optimal and collision-free path from an initial point to the target point in the workspace while avoiding all the static obstacles or other mobile agents as well as taking into account kinematic constraints\cite{yang2016survey}. It has gained popularity among researchers around the world.  

In the last decades, many sampling-based path-planning algorithms have been explored and introduced. Among them, Rapidly Exploring Random Tree (RRT) is one of the quickest and most efficient obstacle-free path-finding algorithms. RRT algorithm was first proposed in \cite{Lavalle98rapidly-exploringrandom} as a sampling way of solving high-dimensional path planning problems. It has been proven to be probabilistically completed, and computationally efficient \cite{lavalle2001rapidly}. 

The advantage of RRT algorithm is that it does not need to model the system or geometrically divide the search space. It has a high coverage in the search space and a wide search scope, so it can explore the unknown space as much as possible. However, the defects of RRT algorithm are also obvious. For example, the algorithm is not deterministic, a narrow passage is also difficult to pass, and the found path is sharp-edged, which can’t be driven easily.

To improve RRT algorithm, many of variants have been proposed in the past 20 years, for example: RRT-connect \cite{kuffner2000rrt}, RRT* \cite{karaman2011sampling}, Bidirectional RRT* \cite{jordan2013optimal}, RRT*-Smart \cite{noreen2016comparison}, SRRT* \cite{hess2016srrt}. These variants improve the efficiency and rate of convergence and solve the problem of asymptotic optimization. However, for robot autonomous control, these pure algorithms are not enough since the dynamic constraints of vehicles are not considered. In \cite{kuwata2009real}, a closed-loop RRT (CL-RRT) is proposed to involve non-holonomic constraints of the four-wheel vehicle dynamics. In \cite{dong2016rrt} the CL-RRT idea was improved and implemented on quadrotor UAVs. Although the UAV non-holonomic constraints and RRT-based controller design method are provided, the Simulation environment is too ideal, and the final path is not smooth enough and not cost-optimal.

\begin{figure}[h] 
	\centering
	\includegraphics[width=0.45\textwidth]{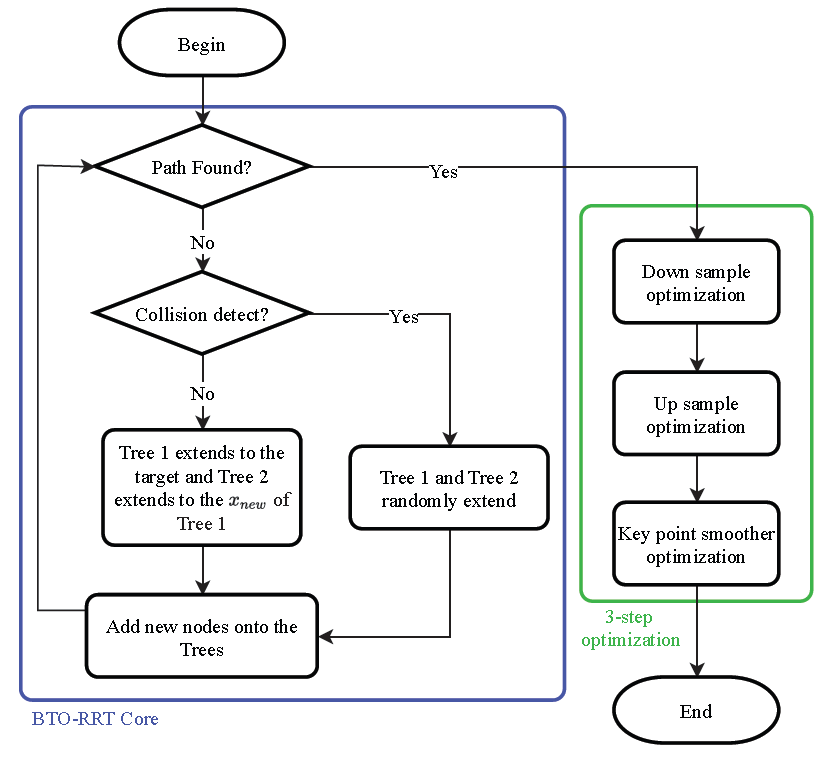}
	\caption[Proposed algorithm flowchart]{Proposed algorithm flowchart}
	\label{fig:flowchat}
\end{figure}

In this paper, we proposed a bidirectional target-oriented RRT(BTO-RRT) based path planning algorithm. The core functions of this algorithm are similar to the one in our previous work \cite{zheng2020point} with some exceptions that will be pointed out later. Taking advantage of the bidirectional RRT algorithm and RRT-connect algorithm, our approach becomes more "target-oriented." The BTO-RRT algorithm core will first generate a zigzagged trajectory. After trajectory generation, the algorithm implements 3-step optimization to shorten further and smooth the total distance of the original path. In the optimization process, we used key point smoother optimization instead of B-Spline Curve, which combines the intermediate point interpolation and cubic-spline curve to guarantee collision-free and continuity in a clustered environment. And this is the major departure from \cite{zheng2020point}. To apply this algorithm in 3D point cloud environments and implement it on ground vehicles and UAVs applications, our algorithm first analyzes the density of the targeted environment, takes the analysis results as the inputs of the BTO-RRT algorithm, and then uses a K-d tree-based obstacle avoidance strategy to avoid obstacles points. This proposed algorithm aims to search for a collision-free, low computational cost, smooth, and dynamically feasible path. 8 different types of 2-d maps and 3 different types of point cloud maps with respective simulations are conducted in this paper to prove the efficiency and generality of this algorithm. 

\section{Bidirectional Target-Oriented RRT-Based Algorithm}
\label{sec:1} 
This paper proposes a bidirectional target-oriented RRT (BTO-RRT) algorithm that searches for obstacle-free, computational low-cost, smooth trajectories. The proposed algorithm contains two parts: one is the BTO-RRT core algorithm, and the other one is the optimization algorithm as you can see from the algorithm flowchart fig. \ref{fig:flowchat}. The core algorithm is to find a more target-oriented initial solution. If the initial path is found, the path will be further improved by the 3-step optimization which includes Down-sample optimization, up-sample optimization and key point smoother optimization. Moreover, with the modification we introduced in the section \ref{sec:2}, this algorithm can be easily applied not only in any 2D maps, but also in 3D point cloud maps.

\subsection{Bidirectional Target-Oriented Algorithm Core}

\begin{figure}[h]
\centering
\begin{subfigure}{.4\textwidth}
\centering
  \includegraphics[width=1.0\linewidth]{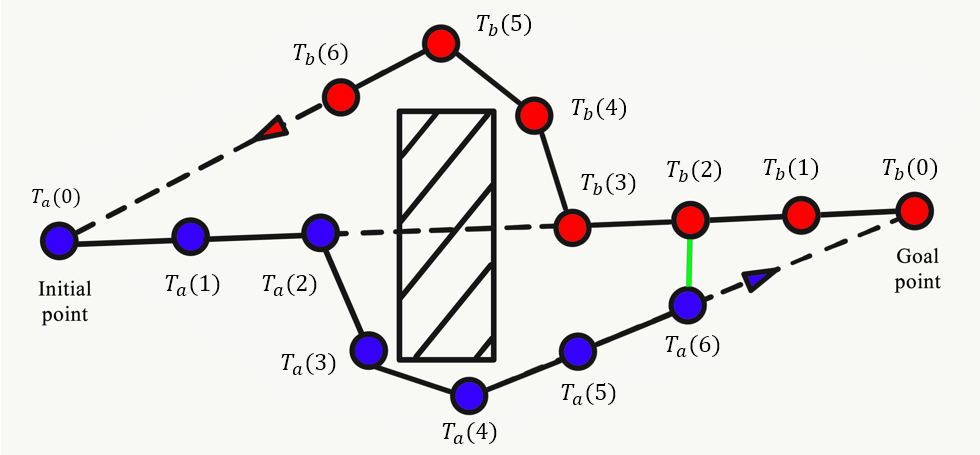} 
  \caption[Bidirectional RRT algorithm]{Bidirectional RRT core algorithm}
  \label{fig:B-RRT-illus}
\end{subfigure}
\begin{subfigure}{.4\textwidth}
\centering
  \includegraphics[width=1.0\linewidth]{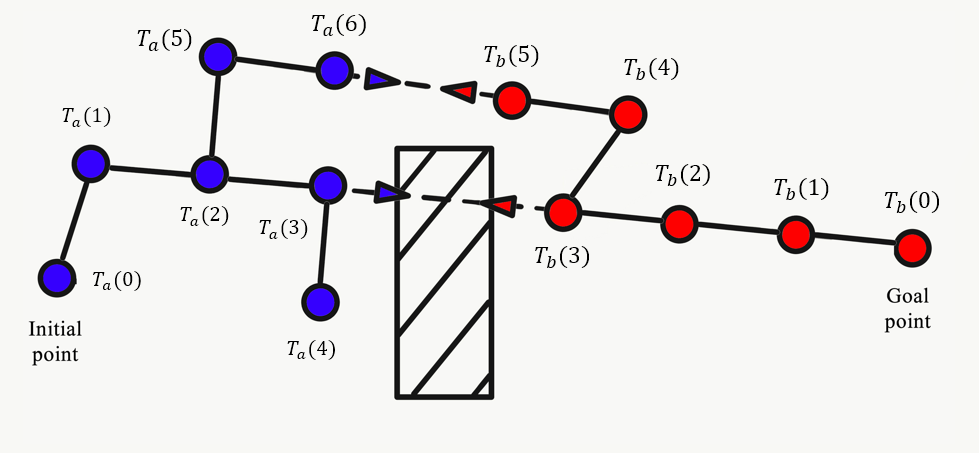} 
  \caption[RRT-Connect algorithm]{RRT-Connect algorithm}
  \label{fig:Connect-RRT-illus}
\end{subfigure}
\begin{subfigure}{.4\textwidth}
\centering
  \includegraphics[width=1.0\linewidth]{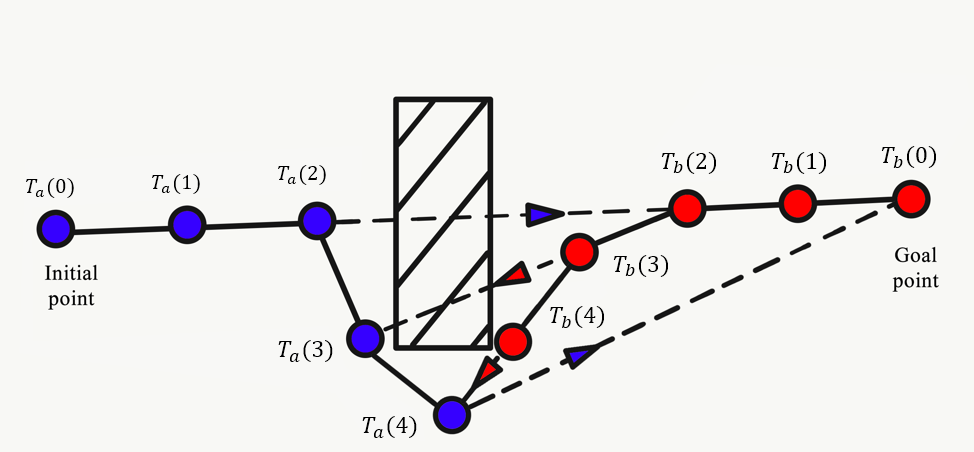} 
  \caption{BTO-RRT algorithm core}
  \label{fig:BTO-RRT-illus}
\end{subfigure}
    \caption{The schematic of the B-RRT, RRT-connect, and  our proposed algorithm core. The blue nodes stand for $T_a$, and the red nodes stand for $T_b$. Different arrows with different colors specify the expansion direction of the tree.}
    \label{fig:3RRT_examples}
\end{figure}

The Bidirectional target-oriented path planning algorithm is a variant of the sample-based algorithm. It takes the advantages of bi-directional RRT(B-RRT), and RRT connect, and it improves the extension rules so it could be more "target-oriented" and connect in a faster way.

In the bi-directional RRT algorithm, two trees that grow from the target point and the initial point are expanded with each other's starting point as the target. If the two trees meet halfway, they will be connected to each other. As you can see from fig. \ref{fig:B-RRT-illus}, $T_a$ expands toward the initial point while $T_b$ expands toward the goal point. Since they all targeted at each other's starting point, they may spend more tree nodes to find the connection point. In the RRT connect algorithm, Two trees that grow from the target point and the initial point respectively expand with the end position of each other as the target, and connect to the last expansion point of each other. As shown in fig. \ref{fig:Connect-RRT-illus}, the end nodes of both $T_a$ and $T_b$ are targeted at each other. This may spend fewer nodes on seeking connections, but it will make the path less "target-oriented". 

\begin{algorithm}
    \caption{\textsc{BTO-RRT}$(x_{init},x_{goal})$} \begin{algorithmic}[1]
    \State $T_a$.init($x_{init}$)
    \State $T_b$.init($x_{goal}$)
    \State $Pathfound \leftarrow False$
    \State $i \leftarrow 0$
    \While{$Pathfound \neq True$ or $i<N$ }
    \State $\{T_a,Flag_a\}$ $\leftarrow$ \textsc{Extend}($T_a,x_{goal}$)
    \State $\{T_b,Flag_b\}$ $\leftarrow$ \textsc{Extend}($T_b,T_a(end)$) 
    \State $Pathfound$ $\leftarrow$ $Pathfound + Flag_b$
    \EndWhile
    \State $Path_a \leftarrow $  \textsc{FindMinimumPath}($T_a$)
    \State $Path_b \leftarrow $  \textsc{FindMinimumPath}($T_b$)
    \State $Path \leftarrow $ \textsc{Concatenate}($Path_a,Path_b$)
    \State \Return{$Path$}
    \end{algorithmic}
\end{algorithm}

\begin{algorithm}
	\caption{\textsc{Extend}$(T, x)$}
	\begin{algorithmic}[1]
		\State \textsc{Flag} $\leftarrow$ 0; $i \leftarrow 0$;
		\If{$\textsc{CollisionDetect} = \textsc{NoCollision}$}
		\State $x_{rand}$ $\leftarrow$ $x$
		\Else 
		\State $x_{rand}$ $\leftarrow$ Sample(i) 
		\EndIf
		\State $x_{nearest}$ $\leftarrow \textsc{NearestNode}(T, x)$ 
		\State $x_{new}$ $\leftarrow$ \textsc{Steer} $(x_{nearest}, x)$
		\If {$\textsc{CollisionDetect} = \textsc{NoCollision}$}
		\State $V \leftarrow$ \{$x_{new}$\}; $E \leftarrow \{x_{new},x_{near}\} $
		\State T $\leftarrow $\{V, E\}
		\If{$|x_{new} - x| < StepSize$}
		\State {$\textbf{return}$ \textsc{Flag} = 1} 
		\EndIf
		\EndIf
		\State $\textbf{return}$ $T,\textsc{Flag}$ 
	\end{algorithmic}
\end{algorithm}

In our algorithm, we define two trees $T_a$ and $T_b$, initial point $x_{init}$, goal point $x_{goal}$, and edge of the tree $E$. Then $T_a : \{x_{init},E\}$ and $T_b : \{x_{goal},E\}$. As can be seen from Fig. \ref{fig:BTO-RRT-illus}, the tree $T_a$ extending from the initial point always treats the goal point as its target, whereas the tree $T_b$ extending from the goal point always treats the most recently added node as the target. In this way, the expansion of the tree is not only more "target-oriented", but also it takes fewer nodes to connect two trees. The pseudocode of the BTO-RRT algorithm code is shown in algorithm 1. In the while loop, $T_a$ targets on the goal point and $T_b$ targets on the end node of $T_b$, the pseudocode of the key function  \textsc{Extend} is shown in algorithm 2. 

\subsection{Algorithm Optimization}

The original path that is generated from the RRT-based algorithm is zigzagged and to meet the requirement of optimality and dynamical feasibility, further optimization is necessary. Building on the zigzagged path, we proposed a 3-step optimization after the initial path is found. The 3-step optimization includes: \begin{itemize}
    \item Down-sample optimization
    \item Up-sample optimization
    \item Key point smoother optimization
\end{itemize}

\subsubsection{Down Sample Optimization}
\label{Downsample}

\begin{figure}[h] 
	\centering
	\includegraphics[width=0.4\textwidth]{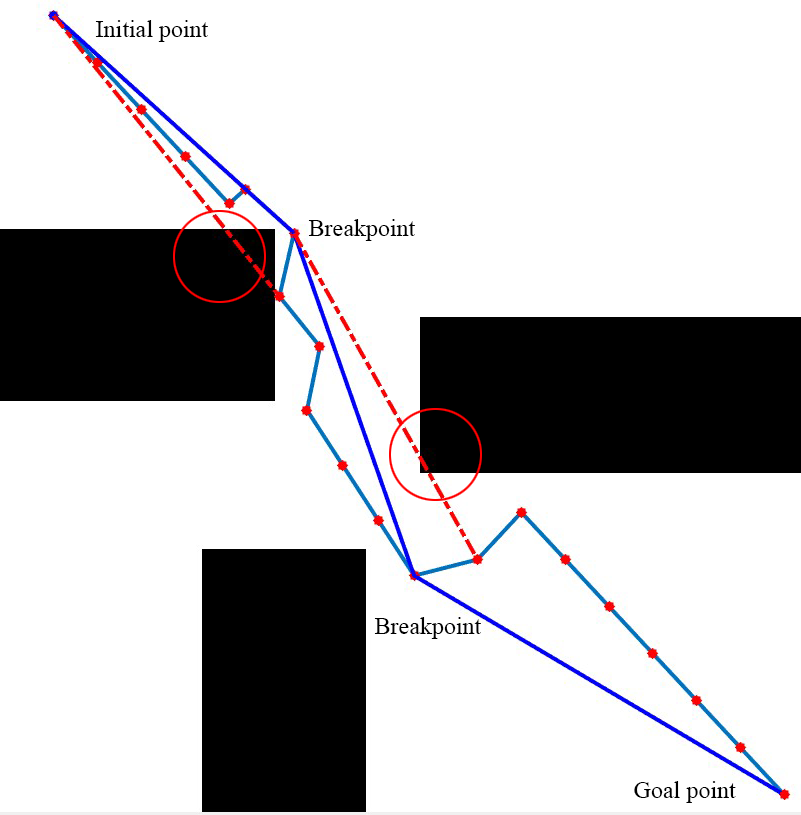}
	\caption[Illustration of down-sample optimization. The blue line is the down-sample path, and the green line-red dotted path is the original path. The red circle is the region where the collision check between two vertices is not true.]{Illustration of down-sample optimization. The blue line is the down-sample path, and the green line-red dotted path is the original path. The red circle is the region where the collision check between two vertices is not true. Therefore, the vertex is discarded and the previous vertex is marked as the breakpoint.}
	\label{fig:downsample_demo}
\end{figure}

Down-sample optimization starts from the initial point and searches along the connection tree to check the collision status between the current point and the next node point. If no collision exists between the current point and its next node point, then our algorithm moves on and keeps checking the collision status between the current point and the point that is behind its next node point. Only after a collision has been detected, the algorithm marks its previous checking node point as the current breakpoint and connects the previous checking point with the last current breakpoint. Essentially, the down-sample optimization is another variant of the greedy algorithm. The illustration of the down-sample optimization is shown in Fig. \ref{fig:downsample_demo}.

\subsubsection{Up Sample Optimization}
\label{Upsample}

\begin{figure*}[h]
\begin{subfigure}{.3\textwidth}
\centering
  \includegraphics[width=1.0 \linewidth]{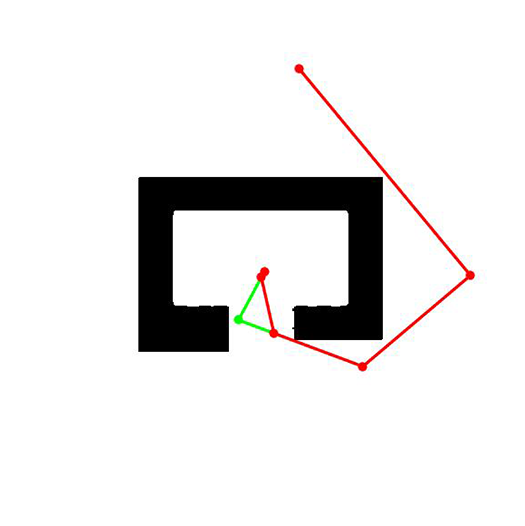}  
  \caption{Iteration time = 10\\path length = 612.44}
  \label{fig:illus_US_a}
\end{subfigure}
\begin{subfigure}{.3\textwidth}
\centering
  \includegraphics[width=1.0 \linewidth]{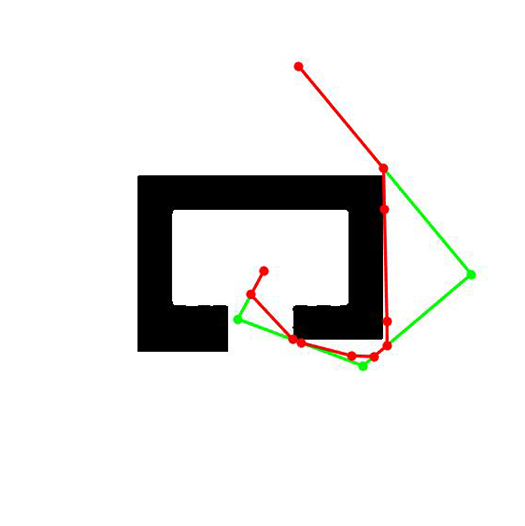}  
  \caption{Iteration time = 100\\path length = 534.94}
  \label{fig:illus_US_b}
\end{subfigure}
\begin{subfigure}{.3\textwidth}
\centering
  \includegraphics[width=1.0 \linewidth]{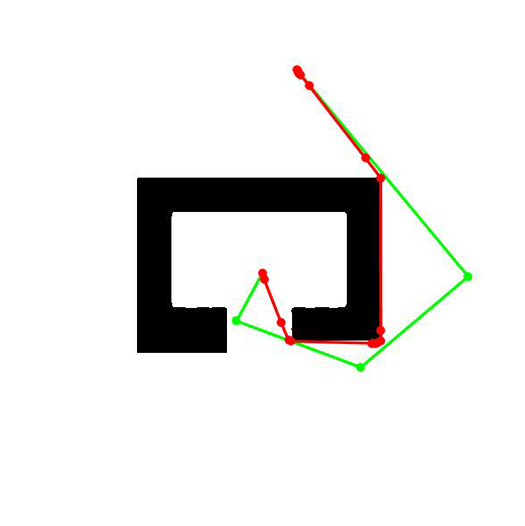}  
  \caption{Iteration time = 1000\\path length = 502.33}
  \label{fig:illus_US_c}
\end{subfigure}
\begin{subfigure}{.33\textwidth}
\centering
  \includegraphics[width=1 \linewidth]{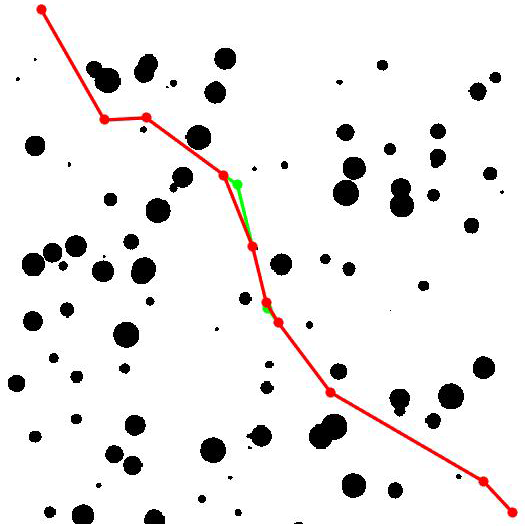}  
  \caption{Iteration time = 10\\path length = 697.13}
  \label{fig:illus_US_d}
\end{subfigure}
\begin{subfigure}{.33\textwidth}
\centering
  \includegraphics[width=1 \linewidth]{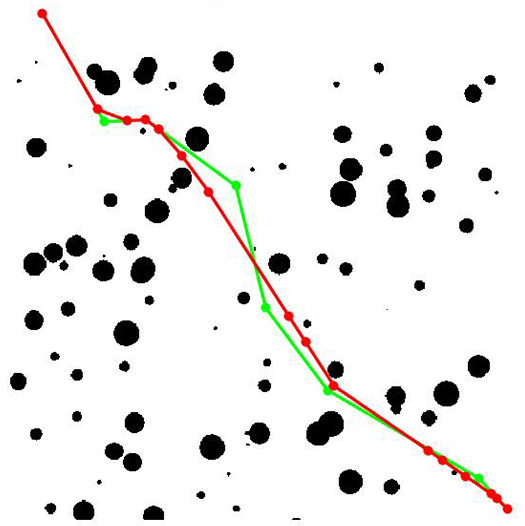}  
  \caption{Iteration time = 100\\path length = 678.88}
  \label{fig:illus_US_e}
\end{subfigure}
\begin{subfigure}{.33\textwidth}
\centering
  \includegraphics[width=1\linewidth]{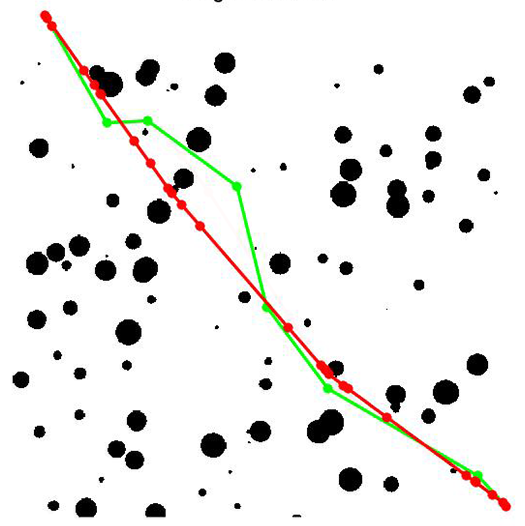}  
  \caption{Iteration time = 1000\\path length = 663.59}
  \label{fig:illus_US_f}
\end{subfigure}
	\caption[The generated trajectories from BTO-RRT core(green lines) and from the up-sample optimization (red lines) with different iteration times]{The generated trajectories from BTO-RRT core(green lines) and from the up-sample optimization (red lines) with different iteration times}
	\label{fig:upsample_illustrate}
\end{figure*} 

After down-sample, the trajectory is locally shorter but it is still based on the zig-zag connected tree, which is not close enough to obstacles especially at the corner. Up-sample optimization is to generate more sample points that are closer to the nearest obstacles and globally shorter which will further shorten the trajectory compared with the down-sample trajectory. 

This up-sample problem can be defined as follows: Consider down-sample trajectory points to be $DST$ and the up-sample trajectory points to be $UST$. The objective is to find a $UST$ that is globally shorter, in which, $DST_n = \{p_1^{DS},p_2^{DS},...,p_n^{DS}\}$, $UST_m = \{p_1^{US},p_2^{US},...,p_m^{US}\}$.  

We will approximate it by sampling iteration. Let $DST = UST$ for the 1st iteration. Define the cumulative sum distance of down sample trajectory points as $Cs$, $Cs_i$ stands for the $i$ element of the $Cs$. Each element will be in the form of:
\begin{equation}
    Cs_i = \sum_{i=1}^{i} p_i^{DS}, \forall i = 1,2,...,n
\end{equation}

Sample two random distance based on the distance between the start point and end point: \begin{equation}
    ranD_{1,2} = x \cdot Cs_{i=n}, x \sim U(0,1)
\end{equation}
where, $ranD_{1,2}$ satisfies: $ranD_1 \leq ranD_2$.

Iteratively insert random interpolation $\gamma_1$ and $\gamma_2$ into the region between $p_i^{US}$ and $p_{j+1}^{US}$, where $i$ satisfies the inequality \ref{equ_i} and $j$ satisfies the inequality \ref{equ_j}, and $\gamma_1$, $\gamma_2$ are defined by Eq.\ref{equ_a1},\ref{equ_a2}, \ref{equ_gamma1} and \ref{equ_gamma2}. The connected path between $\gamma_1$ and $\gamma_2$ should be checked to be collision-free. 
\begin{equation}
\label{equ_i}
     C_{s_i} \leq randD_1 < C_{s_{i+1}}
\end{equation}
\begin{equation}
\label{equ_j}
     C_{s_j} \leq randD_2 < C_{s_{j+1}}
\end{equation}

\begin{equation}
\label{equ_a1}
    a_1 = \frac{ranD_1-Cs_i}{Cs_{i+1}-Cs_{i}}
\end{equation}
\begin{equation}
\label{equ_a2}
    a_2 = \frac{ranD_2-Cs_j}{Cs_{j+1}-Cs_{j}}
\end{equation}
\begin{equation}
\label{equ_gamma1}
    \gamma_1 = (1-a_1) \times p_i^{US} + a_1 \times p_i^{US}
\end{equation}
\begin{equation}
\label{equ_gamma2}
    \gamma_2 = (1-a_2) \times p_i^{US} + a_2 \times p_i^{US}
\end{equation}
Then the new up sample points $UST_{m}^{'}$ becomes:
\begin{equation}
    UST_{m}^{'} = \{ p_1^{US},...,p_i^{US},\gamma_1,\gamma_2,p_{j+1}^{US},...,p_m^{US} \}
\end{equation}

  Let new $DST_n^{'}=UST_m^{'}$ and repeat the above process until it reaches the maximum iteration that we set. By doing so, the vertices from the original path that is globally further will be removed and globally shorter vertices will be added to the path. Since this is a sampling iteration method, the more iterations, the better the performance. Fig. \ref{fig:upsample_illustrate} shows that higher iteration will give us better performance in terms of the distance of up sample trajectory. The sub-figures (a) and (d) are when the iteration number is 10, (b) and (e) is when the iteration number is 100, the (c) and (f) are when the iteration number is 1000. The green lines are the down-sample trajectories and the red lines are the up-sample trajectories with different iteration times. 

  \subsubsection{Key point smoother optimization}

To further smooth the up-sample trajectories, we adapted cubic spline into the optimization process. A cubic spline is a spline constructed of piecewise third-order polynomials which pass through a set of n control points. According to \cite{bartels1995introduction}, a cubic spline $S(x)$ can be defined as: $S(x)$ is a cubic polynomial $S_j(x)$ on [$x_j,x_{j+1}$], $\forall j = 0,1,...,n-1$.
\begin{equation}
    S_j(x) = a_j + b_j(x-x_j) + c_j(x-x_j)^2 + d_j(x-x_j)^3
\end{equation}
\begin{equation}
     h_j = x_{j+1} - x_j, \forall j =0,1,...,n-1
\end{equation}

And it should satisfy the following constraints: \begin{itemize}
    \item In each interval [$x_j,x_{j+1}$], $S(x)$ is given by a cubic polynomial $S_j(x)$.
    \item Interpolation conditions: $S(x_j)=a_j$ for all $j \in {0,1,...,n-1}$
    \item Twice continuously differentiable: For each $j \in {0,1,...,n-1}$, it holds that $S_{j}^{'}(x)=S_{j+1}^{'}(x)$ and $S_{j}^{''}(x)=S_{j+1}^{''}(x)$
    \item At its ends, the curvature of $S(x)$ vanishes: $S_0^{''}(x)=0$,$S_{n-1}^{''}(x)=0$
\end{itemize}

We can Solve for coefficients $a_j,b_j,c_j,d_j$ by using the above constraints: \begin{itemize}
    \item $S_j(x_{j}) = a_j$
    \item $S_{j+1}(x_{j+1}) = a_{j+1} = a_j + b_j h_j + c_j(h_j)^2 + d_j (h_j)^3$
    \item $S^{'}_{j}(x_j) = b_j$, also $b_{j+1} = b_j + 2c_j h_j + 3d_j (h_j)^2$
    \item $S^{''}_{j}(x_j) = 2c_j$, also $c_{j+1} = c_j + 3d_j h_j$
    \item Natural or clamped boundary conditions
\end{itemize}

Therefore, the coefficients $a_j,b_j,c_j,d_j$ are:\begin{itemize}
    \item $a_j = S_j(x_{j})$
    \item $b_j = \frac{a_{j+1}-a_j}{h_j} - \frac{h_j(2c_j+c_{j+1})}{3}$
    \item $c_j$ can be solved in $Ac = B$ format where $A\in \mathbf{R}^{n\times n}, B \in \mathbf{R}^{n\times 1}$,\\ since: $h_{j-1}c_{j-1} + 2(h_{j-1}+h_j)c_j + h_j c_{j+1} = \frac{3}{h_j}(a_{j+1}-a_j) - \frac{3}{h_{j-1}}(a_j - a_{j-1})$ for $\forall j =1,2,...,n-1$
    \item $d_j = \frac{c_{j+1}-c_j}{3h_j}$
\end{itemize}

\begin{algorithm}[h]
    \label{algorithm:keypoint}
	\caption{\textsc{KPSmoothOptimization}($Kpoints,Map$)}
	\begin{algorithmic}[1]
		\State $SPoints$ $\leftarrow$ \textsc{CubicSpline}($Kpoints$)
		\While {\textsc{CheckCollision}($SPoints$)=Collision}
		\State $Kpoints$ $\leftarrow$  \textsc{SmoothOptimizer}($Spoints,Map$)
		\State $SPoints$ $\leftarrow$ \textsc{CubicSpline}($Kpoints$)
		\EndWhile
        \Function{\textsc{SmoothOptimizer}}{$Spoints,Map$}
        \State $MidPoints$ $\leftarrow$ \textsc{CreateMid}($Spoints$)
        \EndFunction
	\end{algorithmic}
\end{algorithm}

\begin{figure}[h]
\begin{subfigure}{.235\textwidth}
\centering
  \includegraphics[width=1.0\linewidth]{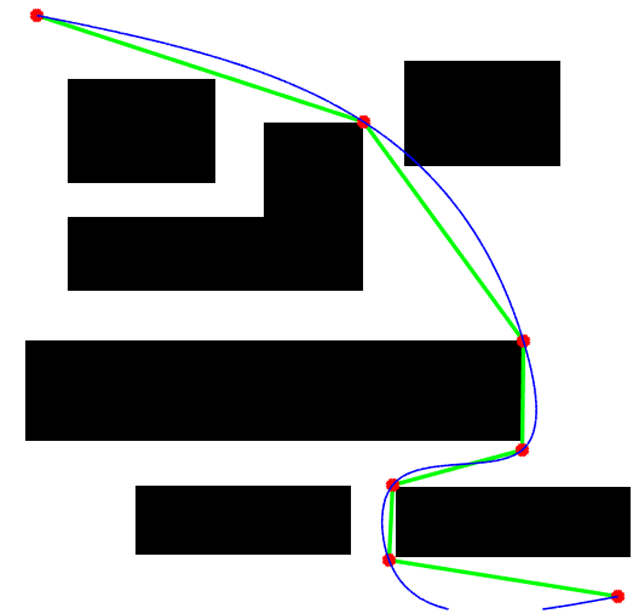}  
  \caption{If implement key point is smoother without checking the collision, the final smooth trajectory will collide with the obstacles in the corner cases}
  \label{fig:upsamplesub-first}
\end{subfigure}
\begin{subfigure}{.235\textwidth}
  \centering
  \includegraphics[width=1.0\linewidth]{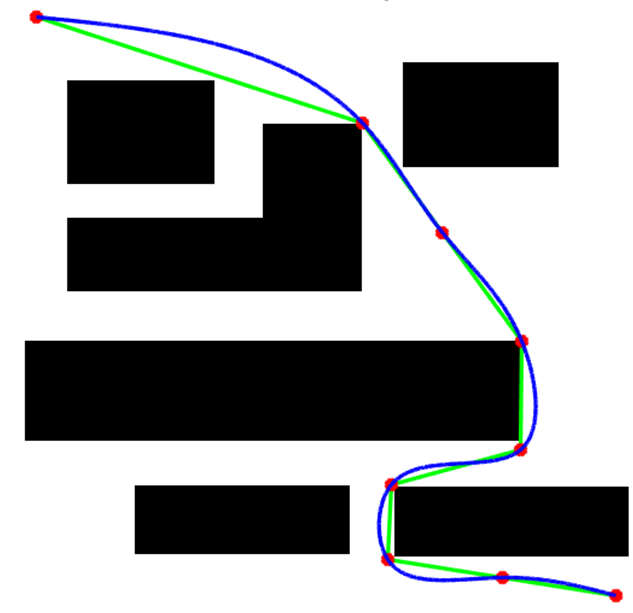}  
  \caption{After interpolation and collision detection in key point smoother optimization, the final smooth trajectory will avoid obstacles while maintaining its continuity}
  \label{fig:upsamplesub-second}
\end{subfigure}
\caption{An illustration figure to show how key point optimization inserts key points to avoid obstacles in a multi-corner map}
\label{fig:fig_supsample}
\end{figure}

In the paper \cite{sprunk2008planning}, Christoph Sprunk proved that cubic spline is second order $C^2$-continuous and it has the ability to visit all the key points (or control points), which will give us the ability to generate a final path that is dynamically feasible and to utilize Key points of interpolation to avoid obstacles. As shown in algorithm 3, Key point smoother optimization takes key points generated by up-sample optimization and maps information as inputs, generating a smooth discretized trajectory $Spoints$. The algorithm first calculates an initial smooth trajectory and then collision checking is done assuming straight lines joining each discrete point along the smooth trajectory. 
If a collision occurs, insert a new key point between the key point closest to the collision point and its previous key point to avoid the collision. This collision detection process will only stop when there is no collision on the final smooth path it will be guaranteed to terminate as long as an up-sample trajectory is found. The effect of conducting a collision check is shown in Fig. \ref{fig:fig_supsample}. 

\section{Point Cloud Based Algorithm and Analysis}
\label{sec:2}
To apply BTO-RRT based algorithm on maps(or environment models), it is important to understand the map environment the algorithm interacts with and adjust the algorithm accordingly. Maps are resources that enable robots to better perform their tasks \cite{rusu2007towards}. Most robot maps are very important and used mainly for robots to accomplish localization and navigation \cite{vasudevan2007cognitive, zhang2021hierarchical, xu2021holistic}. Many robot maps representation has been proposed in the past decades, for example, grid maps, polygonal maps, occupancy maps, counter maps, and 3D mesh maps. In our case, we focus on point cloud maps.

A point cloud is a set of data points in space that contains color and position information of the real world. It is a relatively simple but very powerful and vivid way to represent the real world. In fact, generating these point clouds is now becoming more and more achievable due to cutting-edge technology like 3D reconstruction, SLAM and high-resolution equipment like 3D scanners or stereo cameras. 

To implement the BTO-RRT based path planning algorithm in point cloud maps, the challenge of this problem is that different point clouds have different levels of density. Higher-density point clouds will provide more information for path-planning algorithms, making planning easier in terms of obstacle avoidance. However, the density point cloud will also increase the computational cost exponentially. To generalize this problem, we need to design an algorithm that could adapt our path-planning algorithm to different point clouds with different density levels. Moreover, to address the problem of computational expense, an efficient obstacle avoidance algorithm is also needed.

\subsection{Point Cloud Maps Analysis Algorithm}
Point cloud density is one of the most important properties of point cloud maps. In the point cloud map analysis, point cloud density has to be maintained at an acceptable level for designing a path planning algorithm. 

Since different point cloud maps have different levels of density, implementing a fixed step length RRT-based algorithm will lead to an "Obstacle penetrating error" problem when planning a path in a very sparse point cloud map or lead to "cannot go through the narrow tunnel" problem when planning a path in a very dense point cloud map. 

\begin{figure}[h] 
	\centering
	\includegraphics[width=0.4\textwidth]{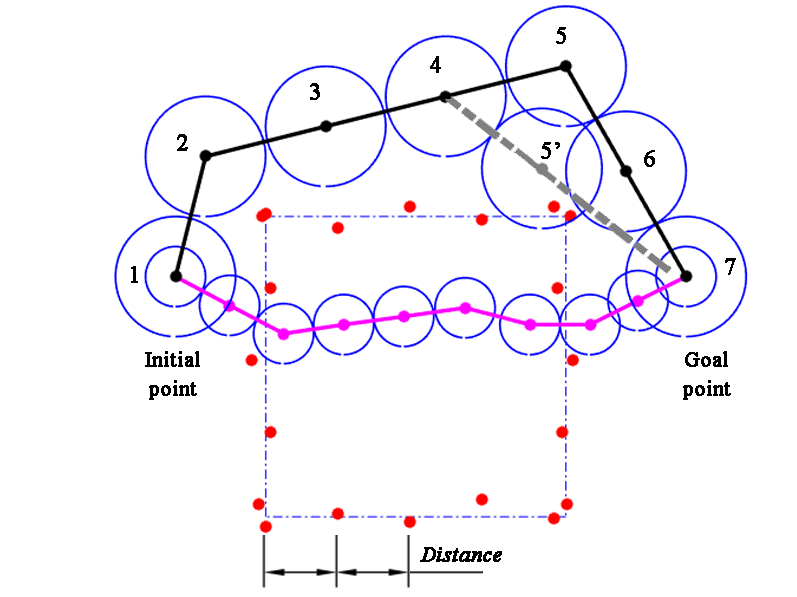}
	\caption[Point cloud analysis and obstacle avoidance ]{Point cloud analysis and obstacle avoidance}
	\label{fig:ptcloud analysis}
\end{figure}

As shown in Fig. \ref{fig:ptcloud analysis}, the red points represent the surface sampling points of the obstacle and the blue dash line represents the shape of the obstacle. If the step length is too small, then the algorithm will generate a magenta path that penetrates the obstacle, which is not what we expected. Therefore, the point cloud analysis algorithm will take the point cloud maps as inputs, analyze the average distance between red sampling points of the obstacle (distance between red dots in the fig. \ref{fig:ptcloud analysis}), and eventually output the step size $Stp$ (the distance between node 1 and node 2 in the fig. \ref{fig:ptcloud analysis}) and safe distance $S$ (radius of the circles in the fig. \ref{fig:ptcloud analysis}). The relationship between $S$ and $Stp$ is shown in the equation \ref{equ1}, where $\alpha$ can be 0.5 or 0.8 : 
\begin{equation}
\label{equ1}
    S = \alpha \times Stp
\end{equation}

\subsection{K-D Tree Based Obstacle Avoidance Algorithm}

\begin{figure*}[h]
\begin{subfigure}{.23\textwidth}
\centering
  \includegraphics[width=1.0\linewidth]{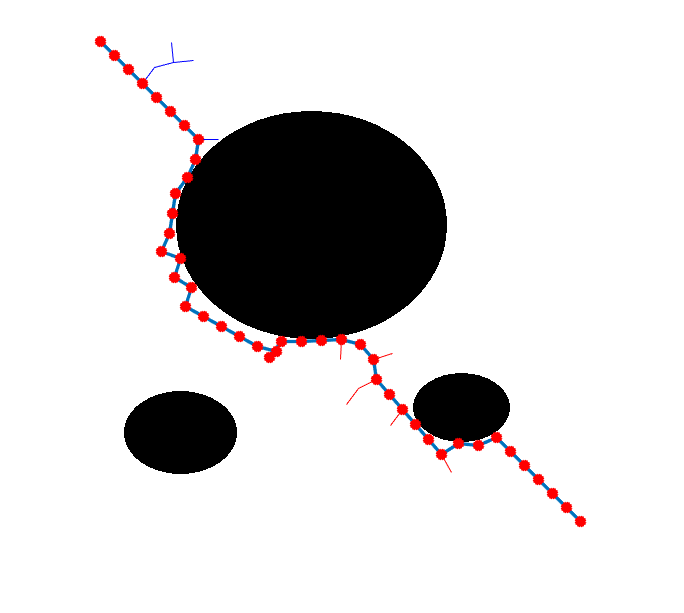}  
  \caption{Map 1 core algorithm result}
  \label{fig:2D-map1a}
\end{subfigure}
\begin{subfigure}{.23\textwidth}
  \centering
  \includegraphics[width=1.0\linewidth]{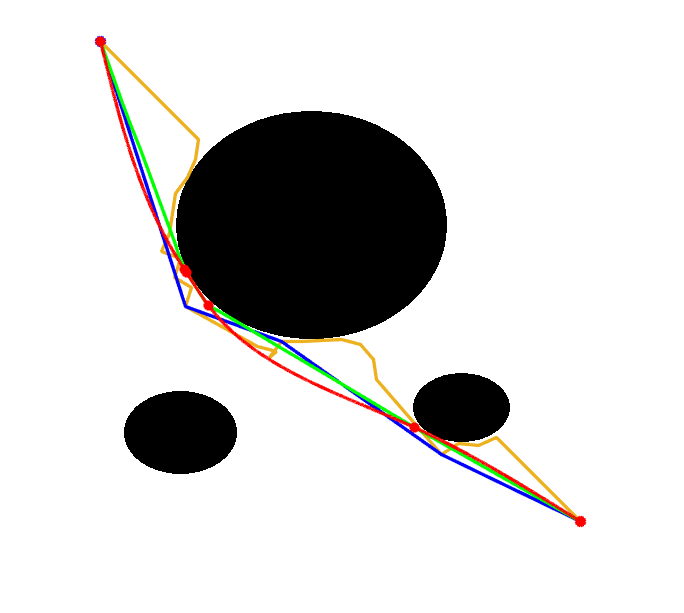}
  \caption{Map 1 optimization result}
  \label{fig:2D-map1b}
\end{subfigure}
\begin{subfigure}{.23\textwidth}
  \centering
  \includegraphics[width=1.0\linewidth]{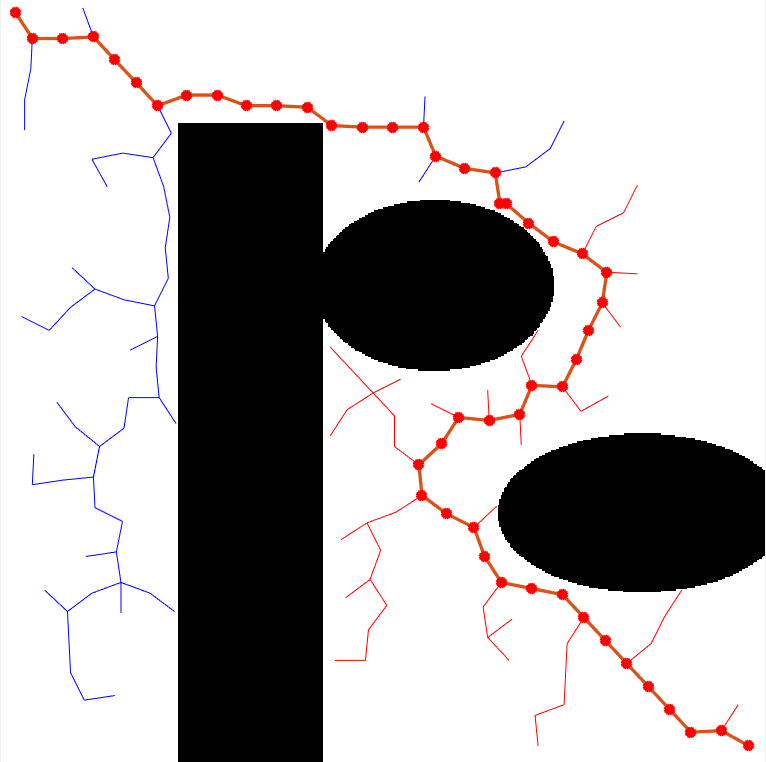}  
  \caption{Map 2 core algorithm result}
  \label{fig:2D-map2a}
\end{subfigure}
\begin{subfigure}{.23\textwidth}
  \centering
  \includegraphics[width=1.0\linewidth]{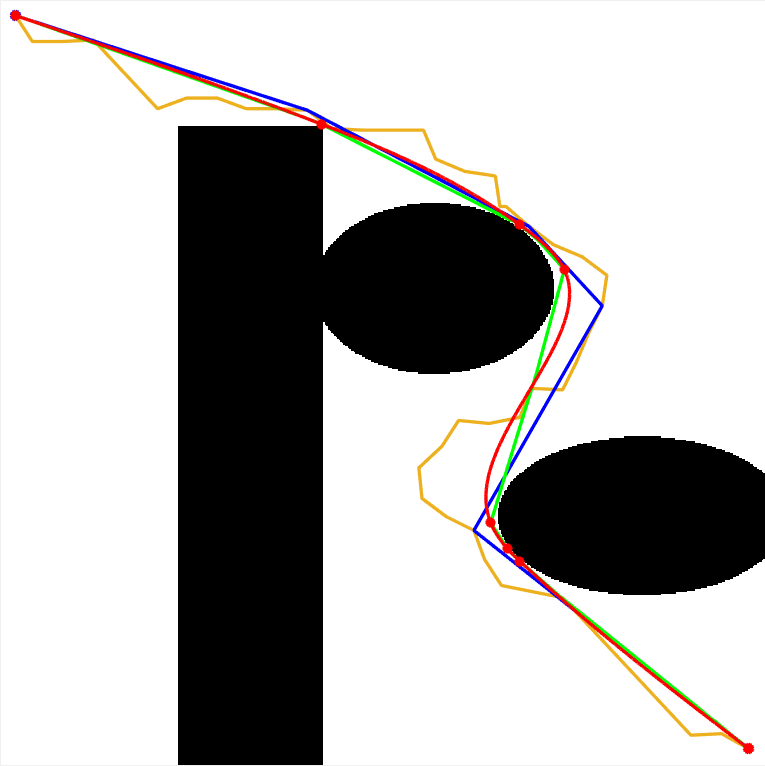}  
  \caption{Map 2 optimization result}
  \label{fig:2D-map2b}
\end{subfigure}
\begin{subfigure}{.23\textwidth}
\centering
  \includegraphics[width=1.0\linewidth]{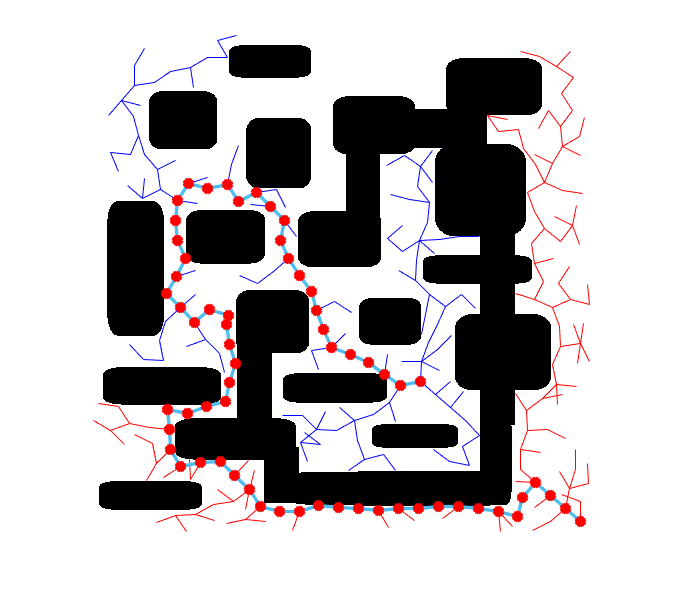}  
  \caption{Map 3 core algorithm result}
  \label{fig:2D-map3a}
\end{subfigure}
\begin{subfigure}{.23\textwidth}
  \centering
  \includegraphics[width=1.0\linewidth]{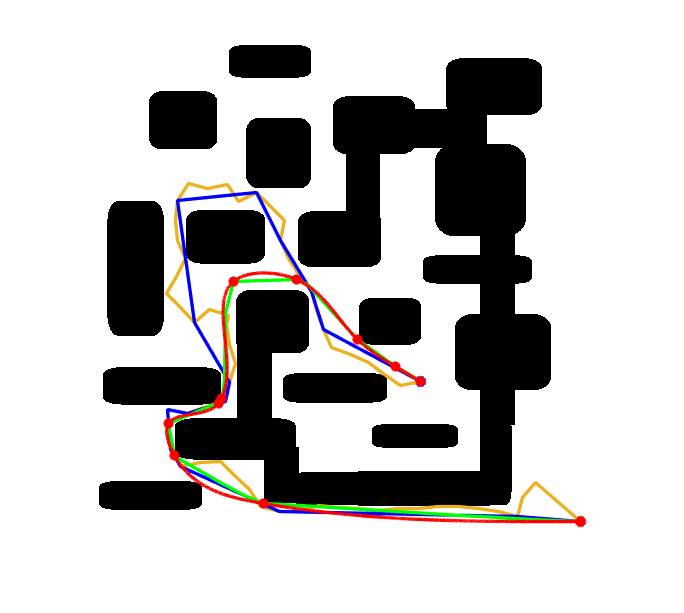}
  \caption{Map 3 optimization result}
  \label{fig:2D-map3b}
\end{subfigure}
\begin{subfigure}{.23\textwidth}
  \centering
  \includegraphics[width=1.0\linewidth]{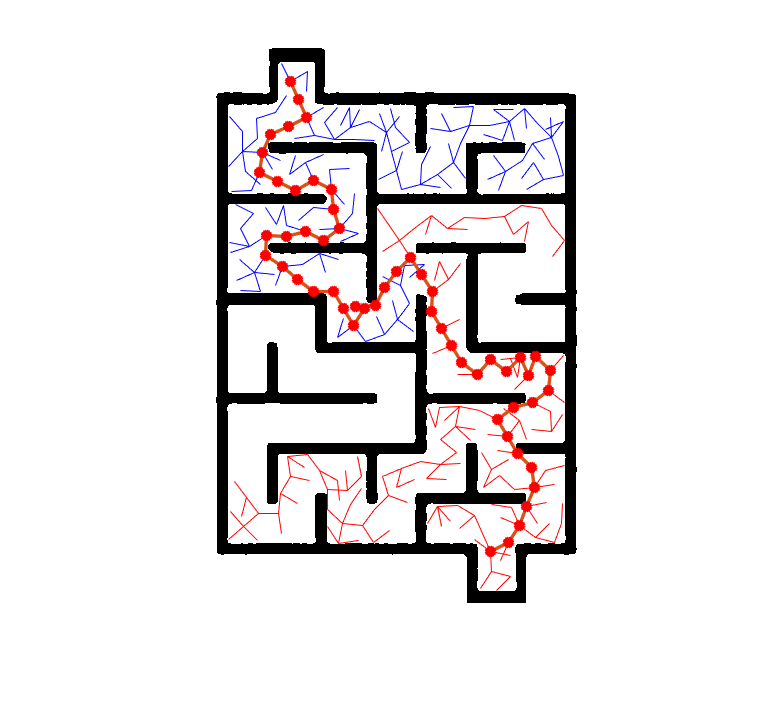}  
  \caption{Map 4 core algorithm result}
  \label{fig:2D-map4a}
\end{subfigure}
\begin{subfigure}{.23\textwidth}
  \centering
  \includegraphics[width=1.0\linewidth]{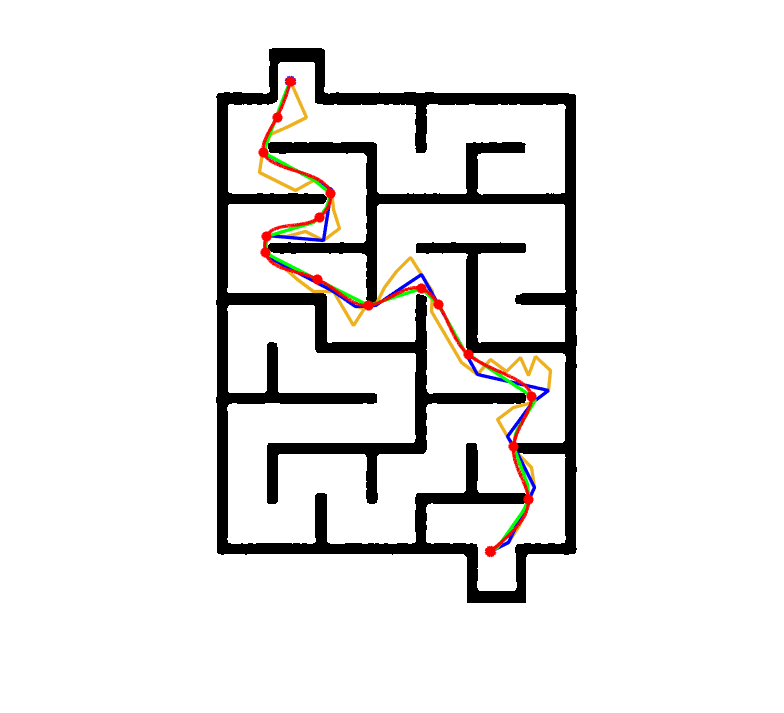}  
  \caption{Map 4 optimization result}
  \label{fig:2D-map4b}
\end{subfigure}
\begin{subfigure}{.23\textwidth}
\centering
  \includegraphics[width=1.0\linewidth]{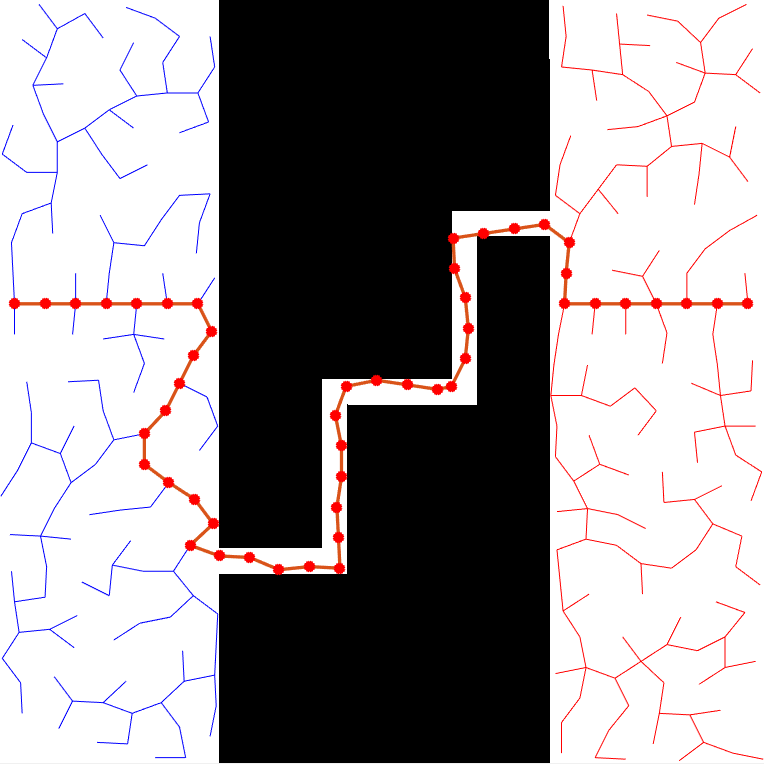}  
  \caption{Map 5 core algorithm result}
  \label{fig:2D-map5a}
\end{subfigure}
\begin{subfigure}{.23\textwidth}
  \centering
  \includegraphics[width=1.0\linewidth]{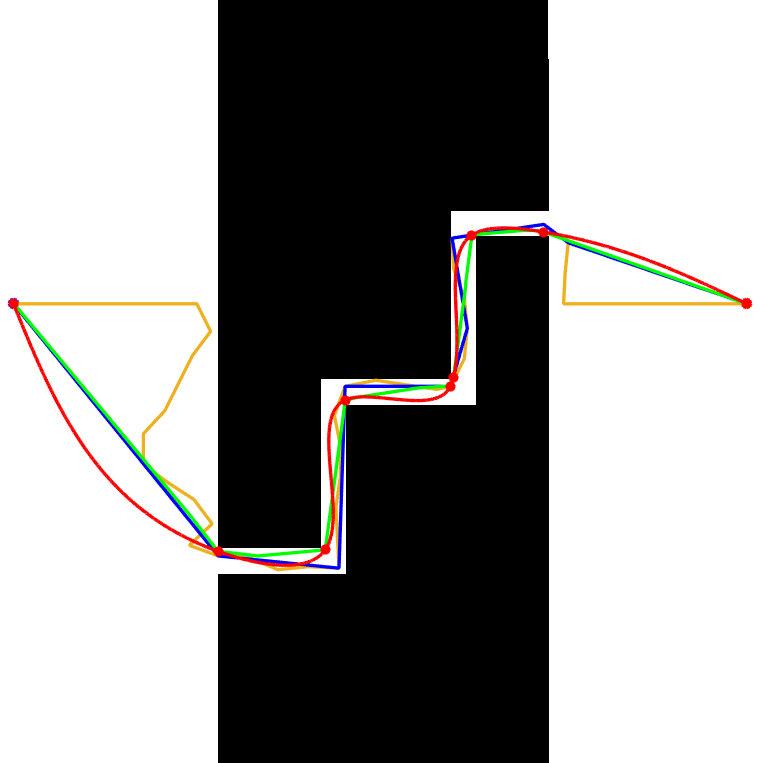}
  \caption{Map 5 optimization result}
  \label{fig:2D-map5b}
\end{subfigure}
\begin{subfigure}{.23\textwidth}
  \centering
  \includegraphics[width=1.0\linewidth]{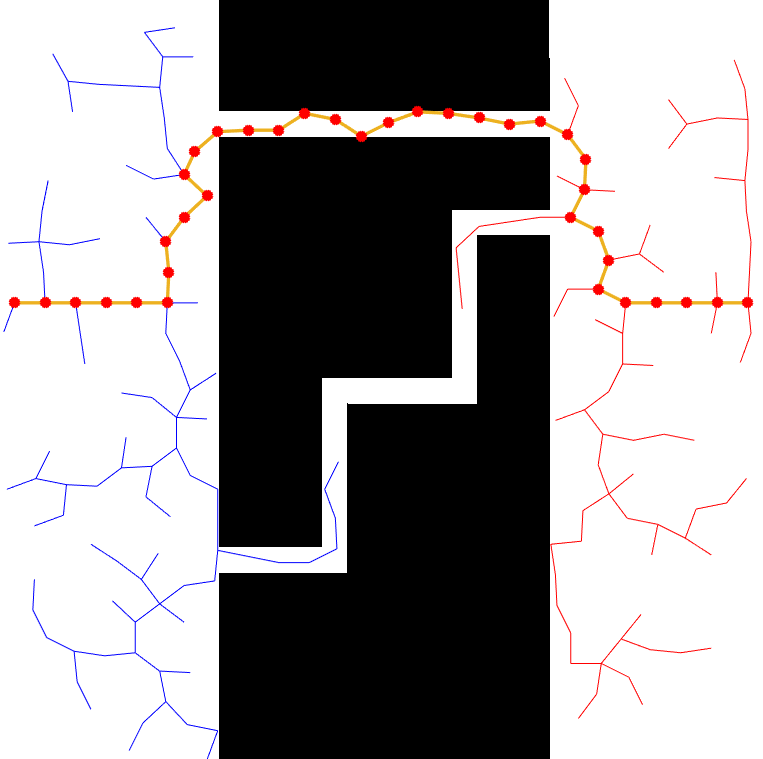}  
  \caption{Map 6 core algorithm result}
  \label{fig:2D-map6a}
\end{subfigure}
\begin{subfigure}{.23\textwidth}
  \centering
  \includegraphics[width=1.0\linewidth]{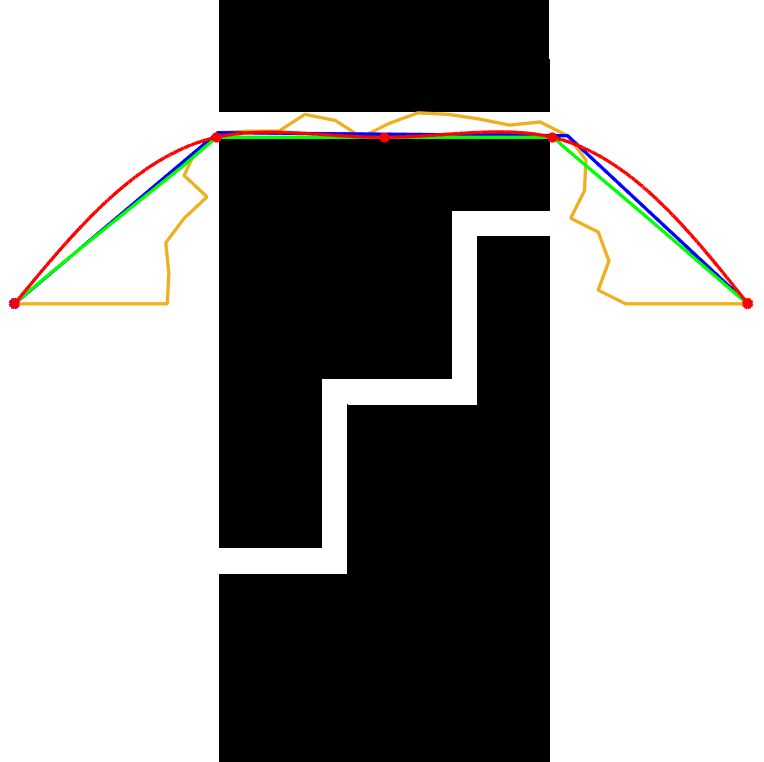}  
  \caption{Map 6 optimization result}
  \label{fig:2D-map6b}
\end{subfigure}
\begin{subfigure}{.245\textwidth}
\centering
  \includegraphics[width=1.0\linewidth]{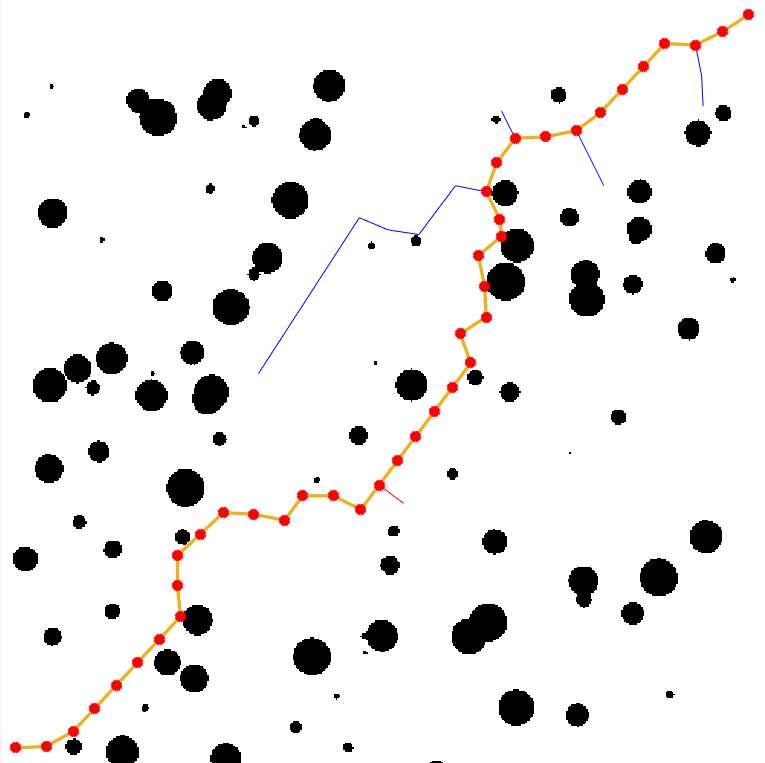}  
  \caption{Map 7 core algorithm result}
  \label{fig:2D-map7a}
\end{subfigure}
\begin{subfigure}{.245\textwidth}
  \centering
  \includegraphics[width=1.0\linewidth]{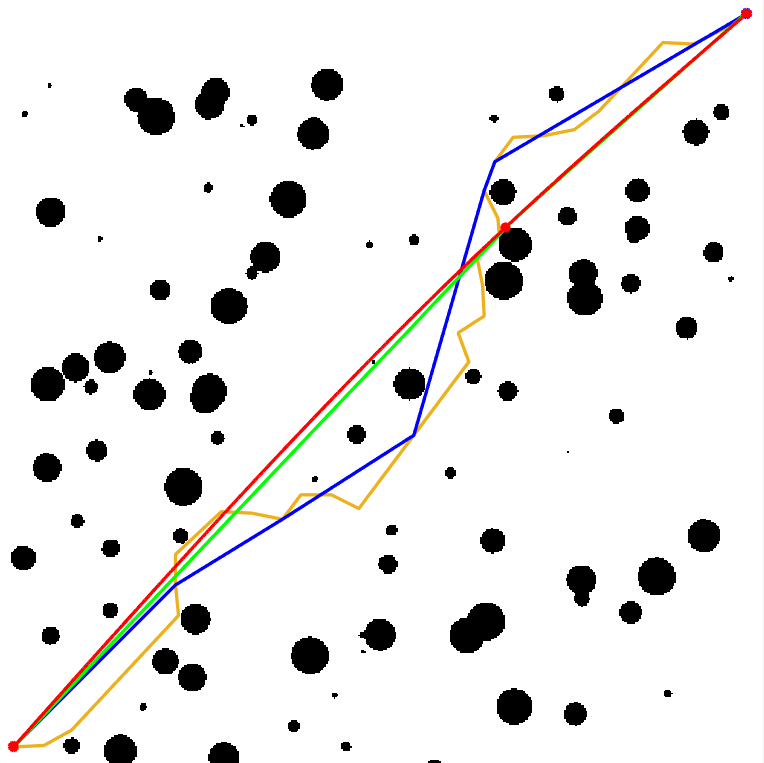}
  \caption{Map 7 optimization result}
  \label{fig:2D-map7b}
\end{subfigure}
\begin{subfigure}{.245\textwidth}
  \centering
  \includegraphics[width=1.0\linewidth]{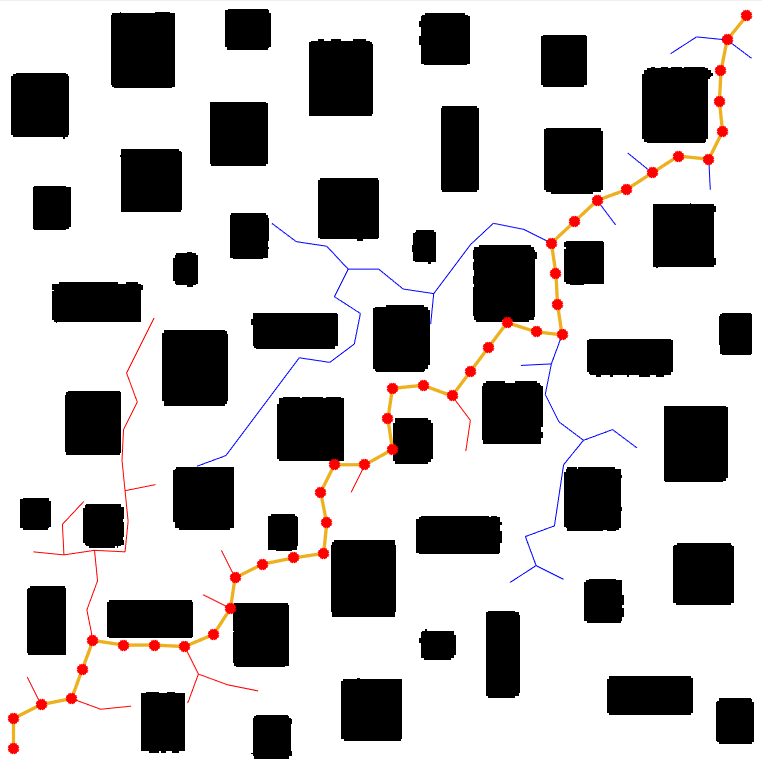}  
  \caption{Map 8 core algorithm result}
  \label{fig:2D-map8a}
\end{subfigure}
\begin{subfigure}{.245\textwidth}
  \centering
  \includegraphics[width=1.0\linewidth]{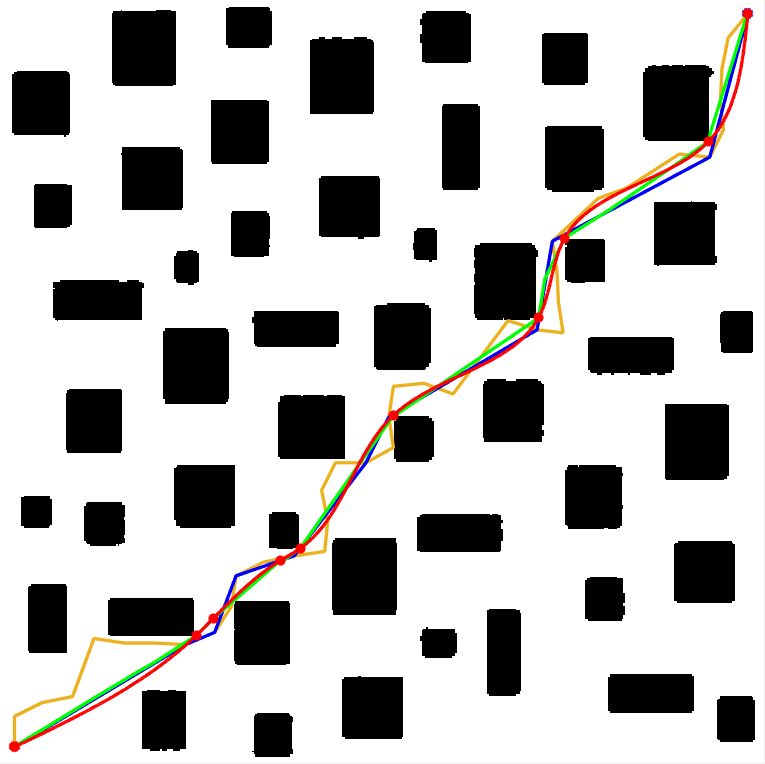}  
  \caption{Map 8 optimization result}
  \label{fig:2D-map8b}
\end{subfigure}
\caption{Performance on different 2D maps}
\label{fig:2D_maps}
\end{figure*}

In  path  planning,  designing  a  feasible  path  is  the  first priority  task.  The  maps  environment  that  most  path  planning  algorithms  are  dealing  with  is  all  full  information geometrical maps like grid maps, occupancy maps, 3D mesh maps,  and  so  on. Under  the assumption that all  the geometry information is known, it is easier to apply an obstacle avoidance strategy.

In a set of sampling data points like point clouds, designing a feasible path will become more complicated. The challenges of path planning in point cloud maps are (1) sparse point clouds will cause the planned path to easily pass through the obstacles represented by the sampling points (2) a large number of sampling points in the point cloud maps will lead to computational expenses. Researchers typically do not deal with point clouds directly. A tensor voting framework is proposed in \cite{gao2016online} to adopt the sampling-based path-finding method to generate a flight corridor with a safety guarantee in 3-D space. Otherwise, a direct way to do global path planning on point cloud maps is to use an extremely dense point cloud as in \cite{fedorenko2018global}. According to the test dataset parameters table in \cite{fedorenko2018global}, its point cloud density is $\geq$ 970 $pts/m^2$. A such point cloud is not just computationally expensive, but also too difficult for SLAM to realize using current technology. Thus, a K-D tree-based collision avoidance strategy is proposed in this paper.

A Kd-Trees is a binary data structure invented in the 1970s by Jon Bentley \cite{bentley1975multidimensional}. The kd-Trees algorithm can be separated into two parts\cite{kakde2005range}: the first part is the algorithm of constructing the kd-tree; the second part is the algorithm of searching for the nearest neighbor. According to the complexity analysis in \cite{kakde2005range}, the time complexity of the kd-trees nearest neighbor algorithm is $O(n^{1-1/d}+k)$, where $d$ is the dimension and k is the k nearest neighbors to retrieve. It is far more efficient than exhaustive $O(n^d)$ and more flexible than oct-trees. 

In Fig. \ref{fig:ptcloud analysis}, node 4 actually ``sees"(there is no obstacle blocking the way) the goal point and tries to expand a  new node $ 5^{'}$ in that direction (gray dashed line). The algorithm will calculate distances between the nearest point cloud neighbors to that node and check if it is in the ball-shape safe region, which is defined by $ S $. Eventually, the node $ 5^{'}$ on the dashed line is discarded and node 5 is randomly selected. Repeat the previous process until the randomly selected node 5 is found which maintains the safe distance from the nearest points in the point cloud, then node 5 is added to the tree.  

\section{Simulation experiment and analysis}
\label{sec:3}
This section is devoted to an experimental study of the algorithms. Two simulations are considered to validate the performance of our algorithm with respect to their running time and the cost of the solution achieved. The validation and analysis of each step of the algorithm optimization will also be described in this section. Both simulation and analysis were implemented in MATLAB 2019a and run on a computer with a 2.4 GHz i5 processor and 16GB RAM running the Windows10 system. 

We considered 8 different 2D maps with different configurations in this simulation. Each map has its own characteristics, and there are similarities between every two sets of maps, which can be compared horizontally. All 8 maps have a size of $500 \times 500$. The experiment settings for each map and their parameters are shown in table \ref{table:2}.

In Fig. \ref{fig:2D_maps}, for each map, two results are presented, one is the result of the BTO-RRT core algorithm, and another is the result of algorithm optimization. These two results are to illustrate the difference between the trajectories generated from the core algorithm and from the optimization. For the core algorithm results, there are three different types of lines, the blue thin line stands for the tree starting from the initial point, the red thin line stands for the tree generated from the goal point, and the red dotted line stands for the found path. For the optimization results, there are 4 different types of lines, the zig-zag orange line is the previously found path, the blue line is the trajectory of the down-sample optimization, the green line is the trajectory of the up-sample optimization, and the red-dotted line is the trajectory of the key point smoother optimization.  

For these 8 maps, we divided them into four groups for horizontal comparison. Map 1 and Map 2 are one group, Map 3 and Map 4 are one group, Map 5 and Map 6 are one group, Map 7 and Map 8 are one group. The first group is to verify the performance of the algorithm when facing circular obstacles of different shapes and positions. The first group is to verify the performance of the algorithm when in irregular mazes and regular mazes. The third group is to verify the performance of the algorithm when faced with narrow pipes of different lengths and shapes. The last group is to verify the algorithm's performance in different shapes and numbers of a clustered environment as can be seen from fig. \ref{fig:2D-map1a} - \ref{fig:2D-map8b}, the core algorithm can find a feasible path, and the optimization steps appear to have shortened the path length on these different 2D environments.

\subsection{2D Map Experiments}

\begin{table}
	\caption{Test Dataset Parameters} 
	\label{table:2} 
\centering
\begin{tabular}{c|cccc}
\hline
Parameter & Start point & Goal point  & obstacle \% & features \\ \hline
map 1      &  (10,10)           &       (490,490)                      &    24.44\%        &   Circular       \\
map 2      &    (10,10)         &       (490,490)                      &    27.31 \%        &   obstacle    \\ \hline
map 3      &    (350,330)         &     (490,490)                        &  39.85\%          &    \multicolumn{1}{c}{\begin{tabular}[c]{@{}c@{}}Irregular\\ mazes\end{tabular}}      \\ 
map 4      &    (50,170)         &      (430,340)                       &   12.95\%         &  \multicolumn{1}{c}{\begin{tabular}[c]{@{}c@{}}Regular\\ mazes\end{tabular}}       \\ \hline
map 5      &    (245,10)         &        (245,490)                     &   40.52\%         &   \multicolumn{1}{c}{\begin{tabular}[c]{@{}c@{}}Single-\\ tunnel\end{tabular}}      \\ 
map 6      &    (245,10)         &      (245,400)                       &   39.04\%         &   \multicolumn{1}{c}{\begin{tabular}[c]{@{}c@{}}Multi-\\ tunnel\end{tabular}}       \\ \hline
map 7      &    (10,490)         &      (490,10)                       &    6.81\%        &  \multicolumn{1}{c}{\begin{tabular}[c]{@{}c@{}}Circular\\ scatter\end{tabular}}       \\ 
map 8      &      (10,490)       &      (490,10)                       &    25.4\%        & \multicolumn{1}{c}{\begin{tabular}[c]{@{}c@{}}Square\\ scatter\end{tabular}}       \\ \hline
\end{tabular}
\end{table}

We considered 8 different 2D maps with different configurations in this simulation. Each map has its own characteristics, and there are similarities between every two sets of maps, which can be compared horizontally. All 8 maps have a size of $500 \times 500$. The experiment settings for each map and their parameters are shown in table \ref{table:2}.

In Fig. \ref{fig:2D_maps}, for each map, two results are presented, one is the result of the BTO-RRT core algorithm, and another is the result of algorithm optimization. These two results are to illustrate the difference between the trajectories generated from the core algorithm and from the optimization. For the core algorithm results, there are three different types of lines, the blue thin line stands for the tree starting from the initial point, the red thin line stands for the tree generated from the goal point, and the red dotted line stands for the found path. For the optimization results, there are 4 different types of lines, the zig-zag orange line is the previously found path, the blue line is the trajectory of the down-sample optimization, the green line is the trajectory of the up-sample optimization, and the red-dotted line is the trajectory of the key point smoother optimization.  

For these 8 maps, we divided them into four groups for horizontal comparison. Map 1 and Map 2 are one group, Map 3 and Map 4 are one group, Map 5 and Map 6 are one group, Map 7 and Map 8 are one group. The first group is to verify the performance of the algorithm when facing circular obstacles of different shapes and positions. The first group is to verify the performance of the algorithm when in irregular mazes and regular mazes. The third group is to verify the performance of the algorithm when faced with narrow pipes of different lengths and shapes. The last group is to verify the algorithm's performance in different shapes and numbers of a clustered environment as can be seen from fig. \ref{fig:2D-map1a} - \ref{fig:2D-map8b}, the core algorithm can find a feasible path, and the optimization steps appear to have shortened the path length on these different 2D environments.

\subsection{Algorithm Analysis, Evaluation and Comparison}


In this section, we systematically analyzed and evaluated the improvement of each step of our algorithm. The focus of the analysis and evaluation  is mainly on the optimization part of the BTO-RRT algorithm. Self-analysis and evaluation are just one way of proving the progress of our proposed algorithm. To prove the efficiency of the BTO-RRT algorithm. We also compared our algorithm with other RRT-based algorithms (RRT, Bi-direction RRT, RRT star) in terms of running time, the number of nodes in the tree, and the cost of the solution achieved. The cost of the solution is defined based on the path length. 

\subsubsection{Down sample optimization analysis}
\begin{figure}[h]
	\centering
	\includegraphics[width=0.4\textwidth]{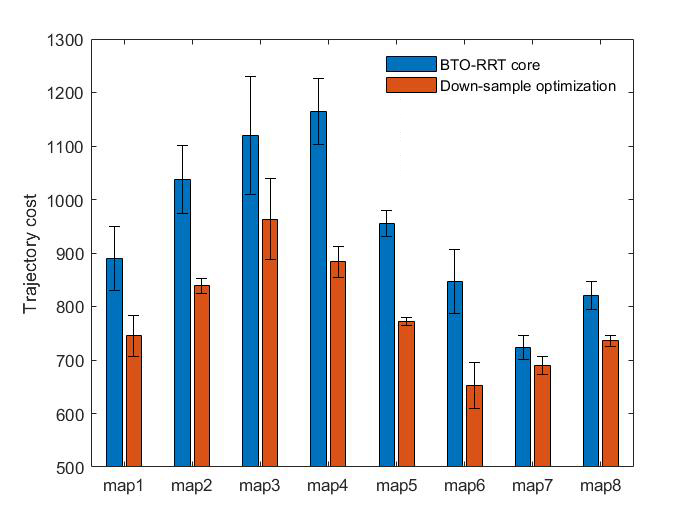}
	\caption[Path length (Cost) comparison before down sample optimization and after down sample optimization]{Path length (Cost) comparison before down sample optimization and after down sample optimization}
	\label{fig:downsample_sim}
\end{figure}
As introduced in the section \ref{Downsample}, the down-sample optimization will shorten the length of the path. To demonstrate that, we ran our algorithm 100 times on 8 maps respectively and statistically calculated the cost of the core algorithm path and the cost of the path after down-sample optimization. As shown in fig. \ref{fig:downsample_sim}, the blue bars stand for the of cost of the path generated by the BTO-RRT core, and the red bars stand for the cost of the path generated after down-sample optimization. The statistical results indicate that down-sample optimization can reduce the cost of the zig-zag core algorithm path by 6 \% - 25 \%. 

\begin{figure}[h]
\begin{subfigure}{0.24\textwidth}
\centering
  \includegraphics[width=1.0\linewidth]{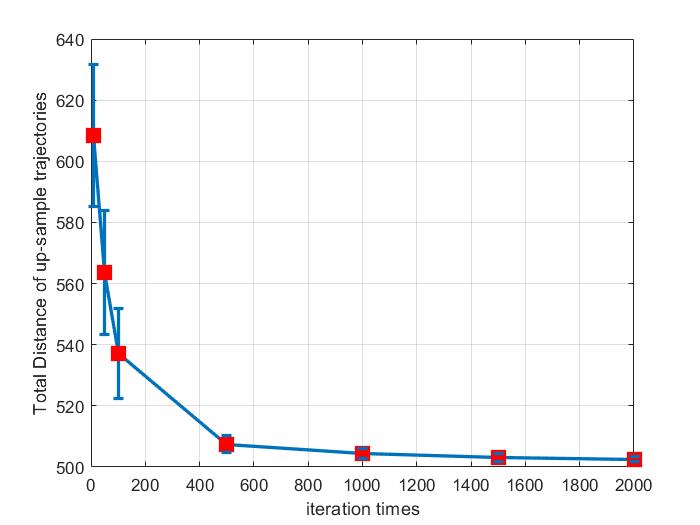}  
  \caption{Map 1 from fig. \ref{fig:upsample_illustrate}}
  \label{fig:upsample-iteration1}
\end{subfigure}
\begin{subfigure}{.24\textwidth}
  \centering
  \includegraphics[width=1.0\linewidth]{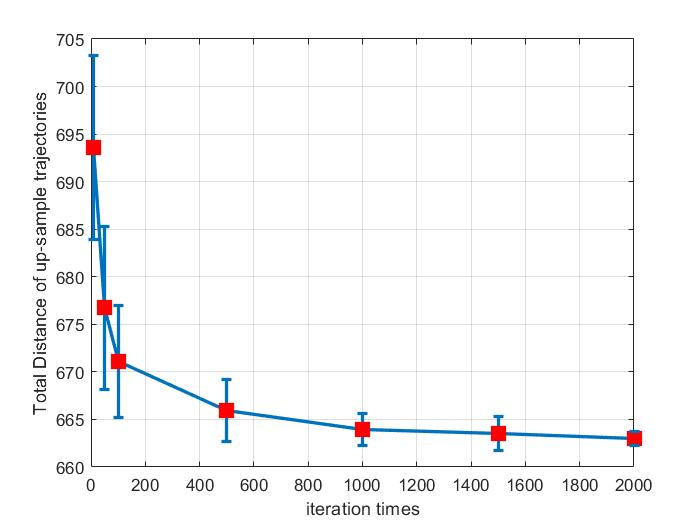}  
  \caption{Map 2 from fig. \ref{fig:upsample_illustrate}}
  \label{fig:upsample-iteration2}
\end{subfigure}
\caption{The trajectory cost of up-sample path over iteration times}
\label{fig:up_sample_sim}
\end{figure}

\subsubsection{Up sample optimization analysis}

\begin{figure}[h] 
	\centering
	\includegraphics[width=0.4\textwidth]{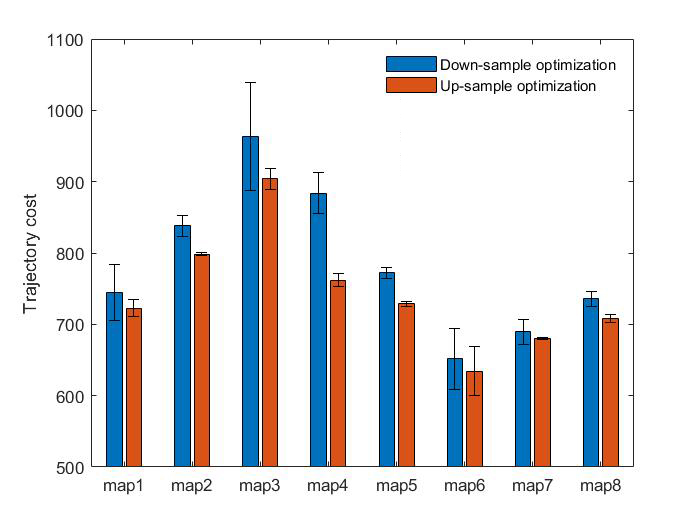}
	\caption[Path length (Cost) comparison before up sample optimization and after up sample optimization]{Path length (Cost) comparison before up sample optimization and after up sample optimization}
	\label{fig:upsample_stat_sim}
\end{figure}

After the down-sample optimization, the path is locally shorter. But this is not an optimal solution. By implementing the up-sample optimization, the cost of the path will approach the asymptotic optimal value, which is the globally optimal solution in the maps. Following up illustration in section \ref{Upsample}, we implemented the up-sample algorithm 100 times on the maps of figure \ref{fig:upsample_illustrate}  and statistically calculated the cost of the path over different iteration numbers as shown in fig. \ref{fig:up_sample_sim}. As the number of iterations increases, the cost of the up-sample path decreases and it will reach an optimal value at some point. From fig. \ref{fig:upsample_stat_sim}, you can clearly see that after up-sample optimization, the cost of paths on all 8 maps is statistically shorter. 

\subsubsection{Key point smoother optimization}
\begin{figure}[h]
\begin{subfigure}{.24\textwidth}
\centering
  \includegraphics[width=1.0\linewidth]{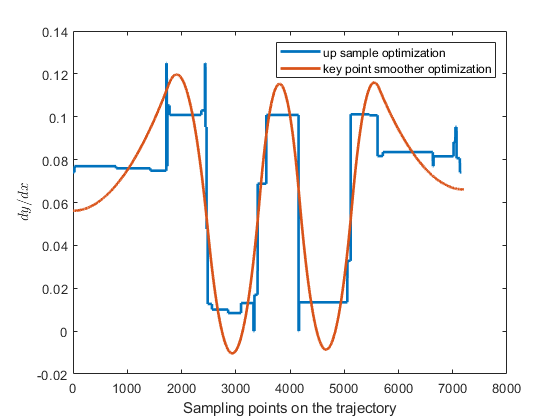}  
  \caption{Derivative curves from fig. \ref{fig:upsamplesub-second}}
  \label{fig:derivative-first}
\end{subfigure}
\begin{subfigure}{.24\textwidth}
  \centering
  \includegraphics[width=1.0\linewidth]{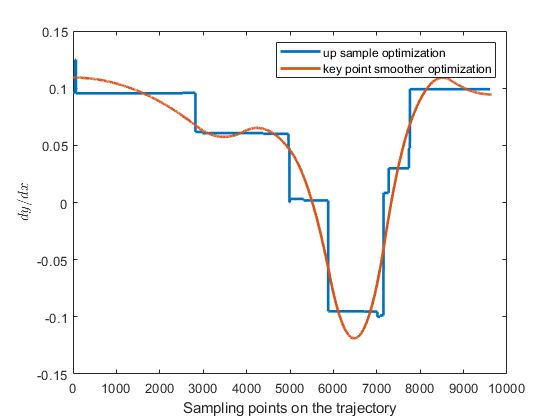}  
  \caption{Derivative curve from fig. \ref{fig:2D-map6b}}
  \label{fig:derivative-second}
\end{subfigure}
\caption{Derivative (dy/dx along the trajectory)curves comparison before and after key point smoother optimization}
\label{fig:key_point_sim}
\end{figure}

After several iterations, the up-sample optimization algorithm converges to a globally shortest optimal solution. However, this optimal path is only trajectory continuous, and it's not dynamically feasible. The key point smoother optimization is to generate a second-order continuous smooth trajectory. In this simulation, we choose two 2D maps (\ref{fig:upsamplesub-second} and \ref{fig:2D-map5b}) as test platforms, and calculate the derivative curve of the up-sample path and key point smoother path. As can be seen from Fig. \ref{fig:key_point_sim}, the red lines and blue lines are the first-order derivative curves of the key point smoother trajectories and the up-sample trajectories. For the up-sampled path, it can be seen from the figure that the inflection point has already appeared in the first derivative. For the key point smoother path, it's a second-order continuous curve, which means that the key-point smoother path is at least velocity continuous.

\subsubsection{Algorithm comparison}
\begin{figure}[h] 
	\centering
	\includegraphics[width=0.4\textwidth]{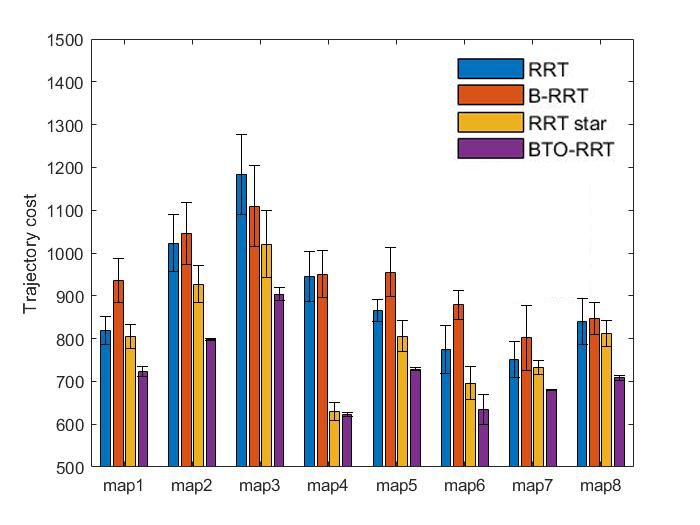}
	\caption[Cost comparison over 8 maps]{Cost comparison over 8 maps}
	\label{fig:cost}
\end{figure}

\begin{figure}[h] 
	\centering
	\includegraphics[width=0.4\textwidth]{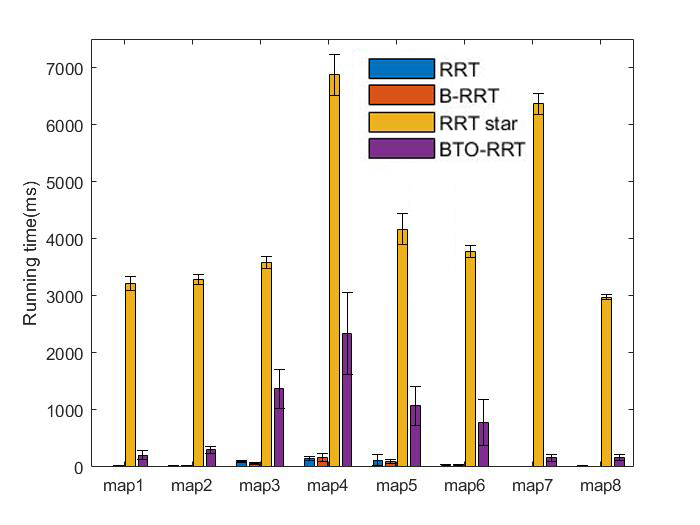}
	\caption[Algorithm running time comparison over 8 maps]{Time comparison over 8 maps}
	\label{fig:time}
\end{figure}

In this section, we compared BTO-RRT algorithm with other RRT-based algorithms like RRT, Bi-RRT, RRT star. Running time and path cost are two very important evaluation criteria of path planning algorithms. In our simulation, we implemented the above four RRT-based algorithms on the previous 8 2d maps and statistically calculate their total running time using the tic/toc function in the MATLAB and cost of the trajectory respectively. All the experiments between different algorithms and all the algorithms themselves are implemented using the same software on the same computer. Since RRT star is an optimal convergence algorithm, and its running time will greatly increase as tree nodes and iteration time become larger and larger, we checked the average data in the literature\cite{karaman2011sampling},\cite{jordan2013optimal},\cite{noreen2016comparison} and chose 4000 as its maximum number of iterations to balance its cost and running time. And since our algorithm will eventually produce a smooth trajectory, which is not being considered in other three algorithms. To better compare the cost without considering the smooth effect, in our cost comparison experiment, we only compared the result from up-sample optimization with the other three algorithms.    

As can be seen from Fig. \ref{fig:cost}, among all algorithms, no matter what type of map, our algorithm can always get the path that has the least cost. Even though RRT star algorithm will eventually converge to the least path cost, from fig. \ref{fig:time}, you can clearly see that even with only 4000 maximum iteration time, the running time of RRT star algorithm is still significantly longer than our BTO-RRT algorithm.

\subsection{3D Point Cloud Map Experiment}

\begin{table}
\caption{3D point cloud maps analysis results and settings for BTO-RRT algorithm}
\centering
\begin{tabular}{cc|p{0.6in} p{0.3in} p{0.3in}}

\hline
\multicolumn{2}{c|}{Parameter}                                                    & \multicolumn{3}{c}{Results from point cloud analysis}                           \\ \hline
Environment                                                                &\multicolumn{1}{c|}{size} & \begin{tabular}[c]{@{}c@{}}Average\\ Density\end{tabular} & Stepsize & SafeDist \\ \hline
\begin{tabular}[c]{@{}c@{}}map1\\ fig.14(a)(d)\\ fig.15(a)(d)\end{tabular} & (63,65,19)      &         0.20 $\pm$ 0.01                                                 & 0.8      & 0.6      \\ \hline
\begin{tabular}[c]{@{}c@{}}map2\\ fig.14(b)(e)\\ fig.15(b)(e)\end{tabular}                                                                       &  (207,207,40)    &    0.34 $\pm$ 0.01                                                        & 2        & 1.5      \\ \hline
\begin{tabular}[c]{@{}c@{}}map3\\ fig.14(c)(f)\\ fig.15(c)(f)\end{tabular}                                                                       &    (270,270,52)  &    0.49  $\pm$ 0.02                                                      & 2        & 1.5      \\ \hline
\end{tabular}

\label{table:3D_map_settings}
\end{table}

\begin{figure*}[h]
\begin{subfigure}{.33\textwidth}
\centering
  \includegraphics[width=.9\linewidth]{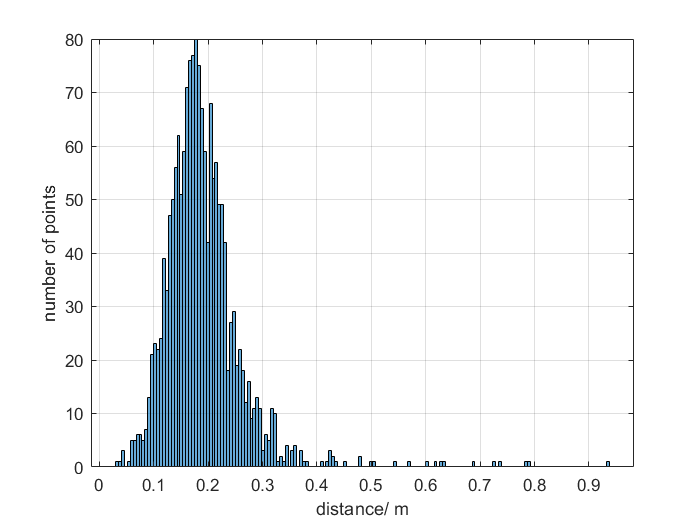}  
  \caption{Point cloud analysis for point cloud map 1}
  \label{fig:SA-map1}
\end{subfigure}
\begin{subfigure}{.33\textwidth}
  \centering
  \includegraphics[width=.9\linewidth]{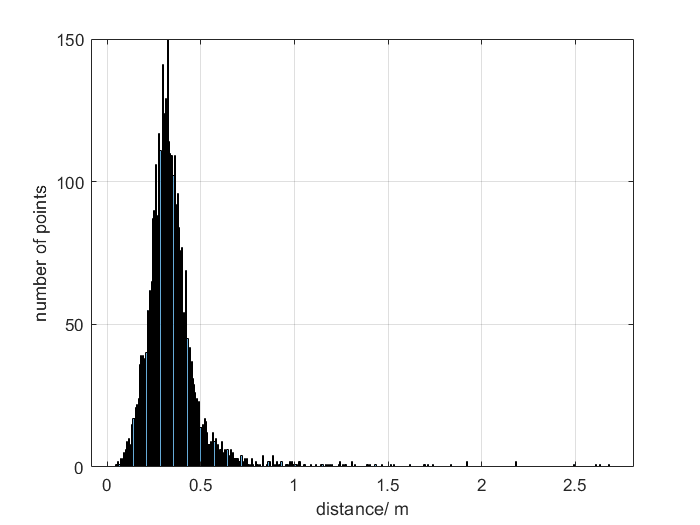}  
  \caption{Point cloud analysis for point cloud map 2}
  \label{fig:SA-map2}
\end{subfigure}
\begin{subfigure}{.33\textwidth}
  \centering
  \includegraphics[width=.9\linewidth]{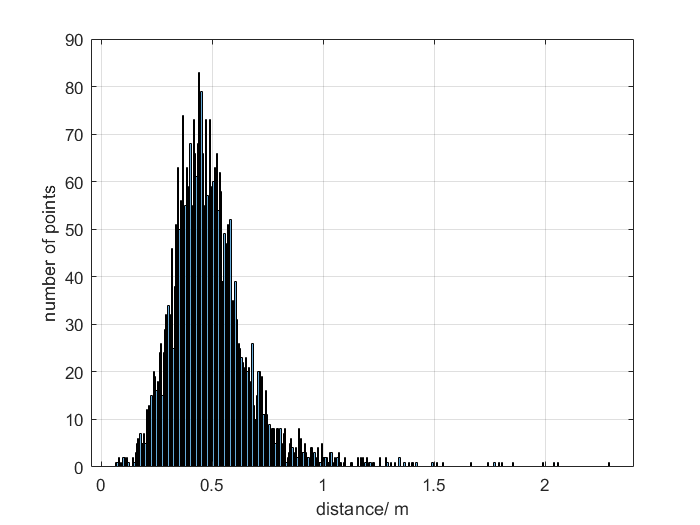}  
  \caption{Point cloud analysis for point cloud map 3}
  \label{fig:SA-map3}
\end{subfigure}
\caption{Point cloud analysis results from different maps}
\label{fig:SA}
\end{figure*}

\begin{figure*}[t]
\begin{subfigure}{.33\textwidth}
\centering
  \includegraphics[width=1.0 \linewidth]{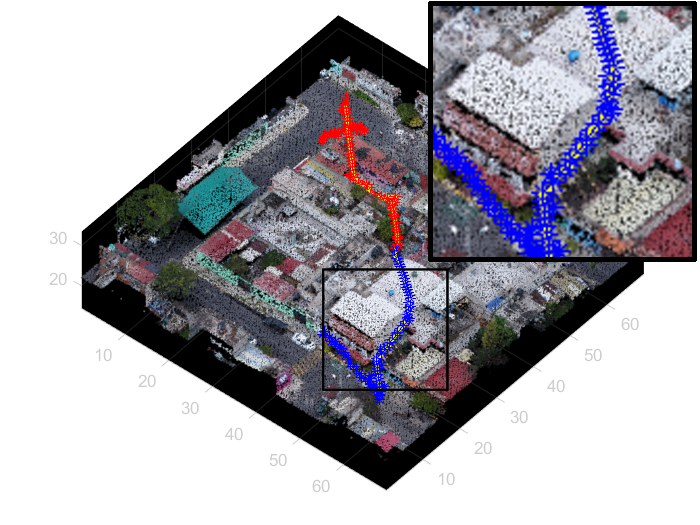}  
  \caption{Map 1 core algorithm result}
  \label{fig:1-3D-map1a}
\end{subfigure}
\begin{subfigure}{.33\textwidth}
  \centering
  \includegraphics[width=1.0\linewidth]{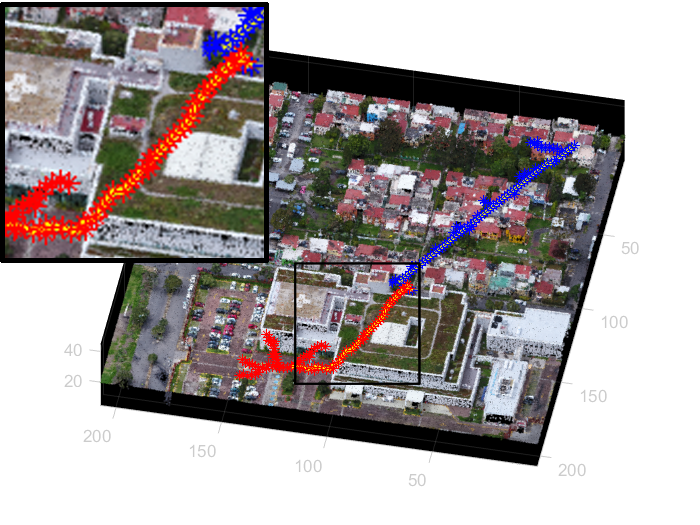}
  \caption{Map 2 core algorithm result}
  \label{fig:1-3D-map2a}
\end{subfigure}
\begin{subfigure}{.33\textwidth}
  \centering
  \includegraphics[width=1.0\linewidth]{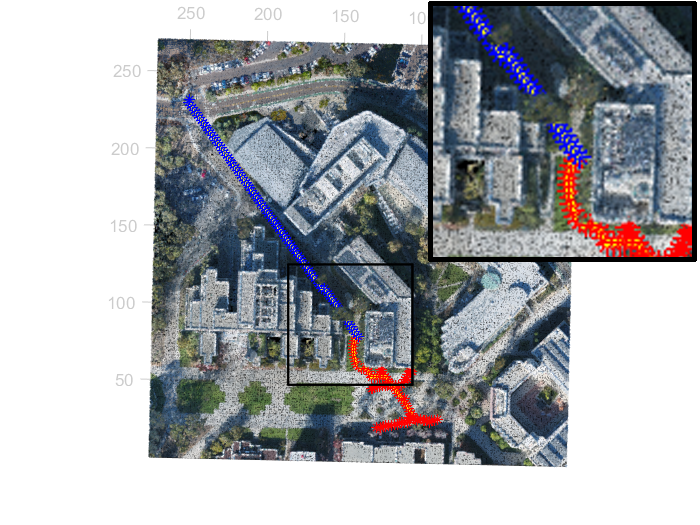}
  \caption{Map 3 core algorithm result}
  \label{fig:1-3D-map3a}
\end{subfigure}
\begin{subfigure}{.33\textwidth}
\centering
  \includegraphics[width=1.0 \linewidth]{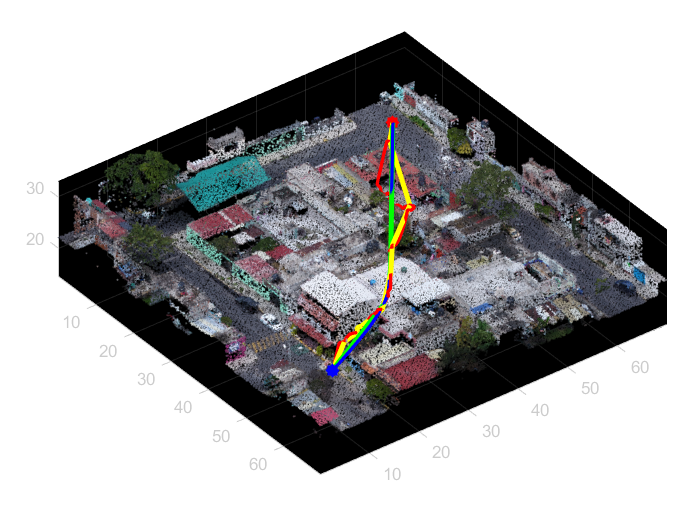}  
  \caption{Map 1 optimization result}
  \label{fig:1-3D-map1b}
\end{subfigure}
\begin{subfigure}{.33\textwidth}
  \centering
  \includegraphics[width=1.0\linewidth]{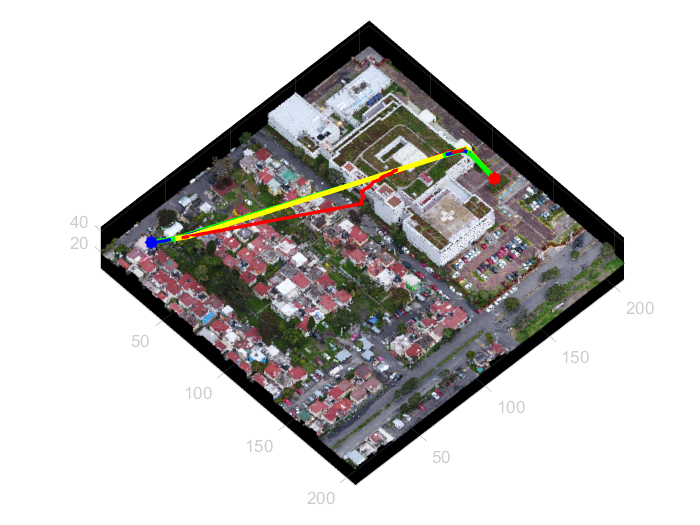}
  \caption{Map 2 optimization result}
  \label{fig:1-3D-map2b}
\end{subfigure}
\begin{subfigure}{.33\textwidth}
  \centering
  \includegraphics[width=1.0\linewidth]{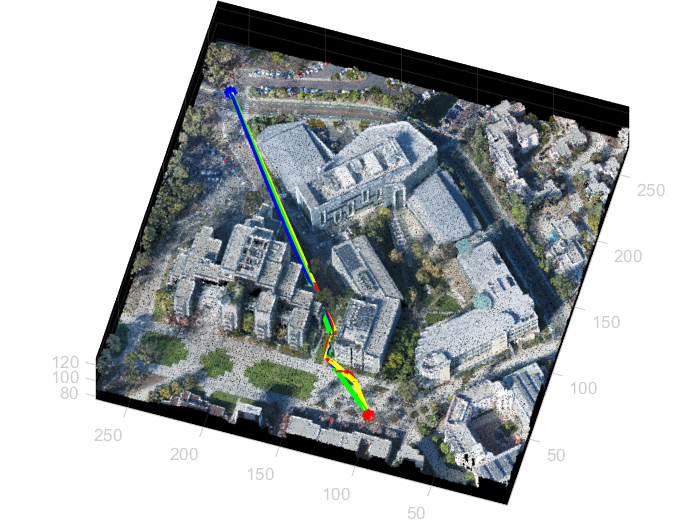}
  \caption{Map 3 optimization result}
  \label{fig:1-3D-map3b}
\end{subfigure}
\caption{$1^{st}$ experiment on 3 different point cloud maps. In fig. (a)(b)(c), the blue trees are generated from initial points and the red trees are generated from goal points. In fig. (d)(e)(f), red lines are the original trajectories, yellow lines are the down-sample trajectories, green lines are the up-sample trajectories, and blue lines are the k-point smooth trajectories.}
\label{fig:3D_maps_1}
\end{figure*}

\begin{figure*}[t]
\begin{subfigure}{.33\textwidth}
\centering
  \includegraphics[width=1.0 \linewidth]{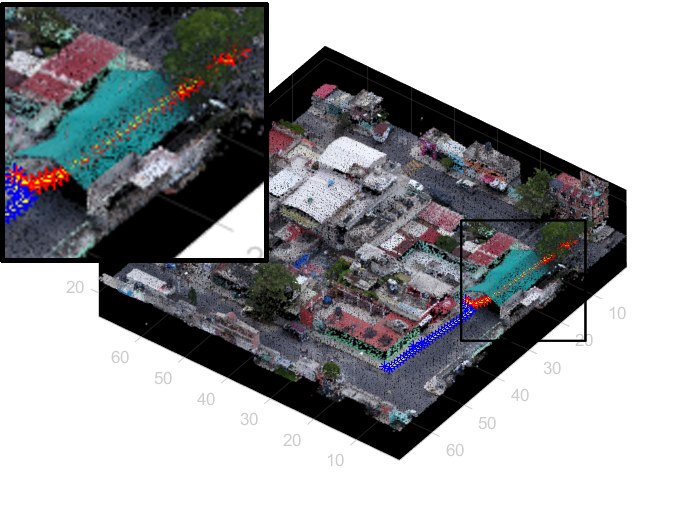}  
  \caption{Map 1 core algorithm result}
  \label{fig:2-3D-map1a}
\end{subfigure}
\begin{subfigure}{.33\textwidth}
  \centering
  \includegraphics[width=1.0\linewidth]{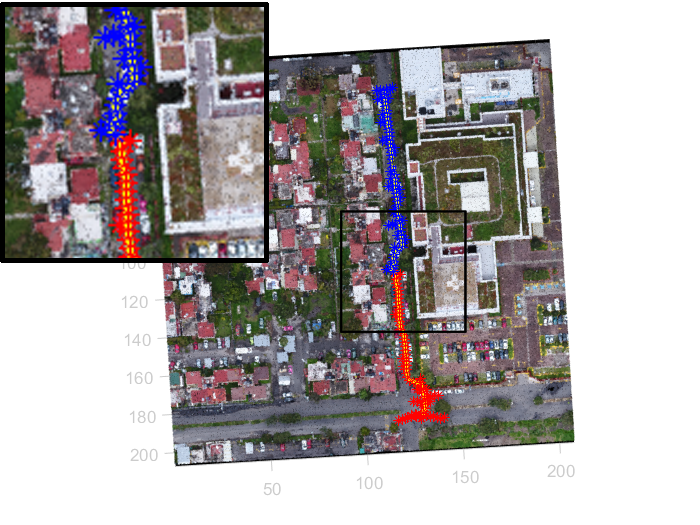}
  \caption{Map 2 core algorithm result}
  \label{fig:2-3D-map2a}
\end{subfigure}
\begin{subfigure}{.33\textwidth}
  \centering
  \includegraphics[width=1.0\linewidth]{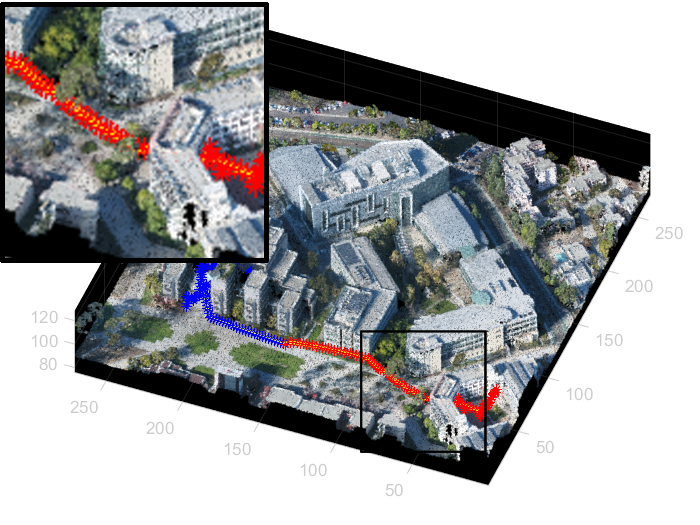}
  \caption{Map 3 core algorithm result}
  \label{fig:2-3D-map3a}
\end{subfigure}
\begin{subfigure}{.33\textwidth}
\centering
  \includegraphics[width=1.0 \linewidth]{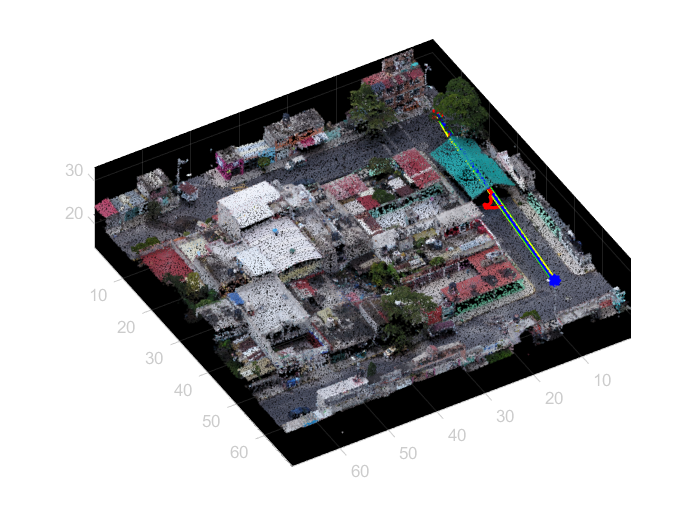}  
  \caption{Map 1 optimization result}
  \label{fig:2-3D-map1b}
\end{subfigure}
\begin{subfigure}{.33\textwidth}
  \centering
  \includegraphics[width=1.0\linewidth]{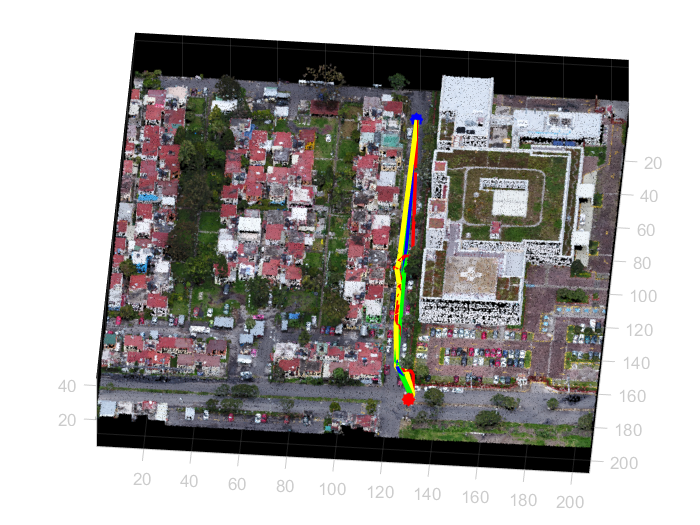}
  \caption{Map 2 optimization result}
  \label{fig:2-3D-map2b}
\end{subfigure}
\begin{subfigure}{.33\textwidth}
  \centering
  \includegraphics[width=1.0\linewidth]{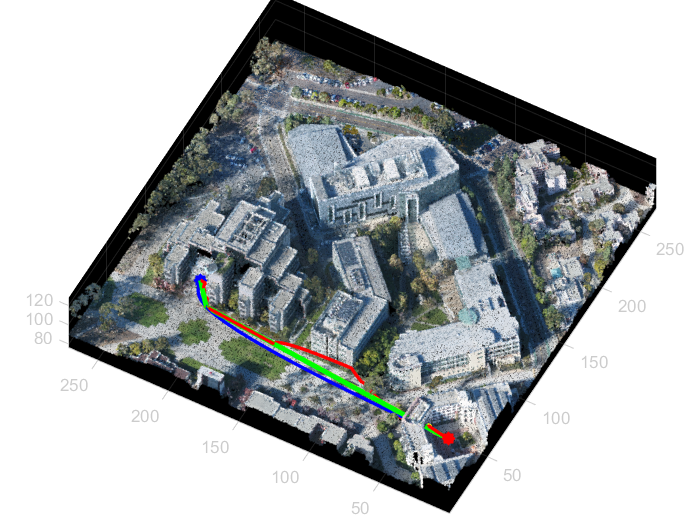}
  \caption{Map 3 optimization result}
  \label{fig:2-3D-map3b}
\end{subfigure}
\caption{$2^{nd}$ experiment on 3 different point cloud maps. In fig. (a)(b)(c), the blue trees are generated from initial points and the red trees are generated from goal points. In fig. (d)(e)(f), red lines are the original trajectories, yellow lines are the down-sample trajectories, green lines are the up-sample trajectories, and blue lines are the k-point smooth trajectories.}
\label{fig:3D_maps_2}
\end{figure*}

To verify the capability of directly utilizing the point cloud information, we consider 3 different point cloud maps with different levels of density. The first 3D map is a residential area, the second 3D map is a hospital area, and the third one is the Jacob school of engineering at UCSD. In the 3D map experiment, we will first verify the point cloud analysis algorithm since this is the first step of implementing BTO-RRT algorithm on a 3D point cloud environment. The inputs of the point cloud analysis algorithm are 3 different point cloud maps, and the outputs are their statistical density analysis shown in fig. \ref{fig:SA} as well as the step size $ Stps $ and safe distance $ S $, which are the inputs of the BTO-RRT algorithm. The results of the point cloud analysis are in table \ref{table:3D_map_settings}. 

We designed two different types of experiments in each 3D map: as can be seen in fig. \ref{fig:3D_maps_1}, the first one is to show how our algorithm bypasses or goes through high buildings; and the second one is to show how the path calculated by the algorithm passes through a narrow corridor as shown in fig. \ref{fig:3D_maps_2}. 

In the first experiment, the initial and goal points we selected are all in the middle of several buildings. This selection is to test and see if our algorithm knows how to cross a building or find a path through a building. In the first 3D map (fig. \ref{fig:1-3D-map1a} and \ref{fig:1-3D-map1b}), the initial point is behind the tallest building of that area. In the second 3D map (fig. \ref{fig:1-3D-map2a} and \ref{fig:1-3D-map2b}), the goal point is in the courtyard of the hospital. In the third 3D map (fig. \ref{fig:1-3D-map3a} and \ref{fig:1-3D-map3b}), the initial and goal point are both behind some very tall buildings. 

In the second experiment, the initial points and goal points we selected are at the end of a narrow tunnel. We want to test to see if the algorithm has the proper step size and threshold to pass these narrow tunnels. In the first 3D map (fig. \ref{fig:2-3D-map1a} and \ref{fig:2-3D-map1b}), the initial point and goal point are at both ends of the green shed. In the second 3D map (fig. \ref{fig:2-3D-map2a} and \ref{fig:2-3D-map2b}), the initial point and goal point are at both ends of the road with lots of trees on the side. In the second 3D map (fig. \ref{fig:2-3D-map3a} and \ref{fig:2-3D-map3b}), the initial point and goal point both are at the courtyard of two buildings. 

In the above experiments, the algorithm demonstrates the ability to select a proper step size/safe distance in different point cloud maps. By doing so, in the first experiment, the generated trajectories bypass the tall buildings instead of penetrating an obstacle. And in the second experiment, the trajectories pass through some narrow corridors instead of bypassing the corridors, proving that selecting step size/safe distance works pretty well in the same point cloud maps under different scenarios. In the optimization process, the down-sample, up-sample and key point smooth optimization can be generalized from 2D to 3D by changing the collision checking algorithm from a 2D version to a 3D K-D tree-based obstacle avoidance algorithm and by changing 2D path coordinate to 3D path coordinate, which proves the generalization of our algorithm. 

\section{Conclusion and Future work}
In this paper, we proposed a general RRT-based path-planning algorithm that could rapidly search for optimal, smooth and dynamically feasible trajectories. The proposed algorithm combined the advantages between bidirectional RRT and RRT-connect to generate a more "target-oriented" initial path, and with 3-step optimization, the initial path is further shortened and converges to a globally shortest smooth path. By utilizing the point cloud analysis and K-D tree-based obstacle avoidance strategy, our algorithm could be successfully implemented on point cloud maps. We believe that by modifying some parameters of the BTO-RRT algorithm, it can also be applied to other types of maps, for example, OctoMaps, occupancy maps, pixel maps and so on. With the capability of rapidly exploring space and generating a smooth path, it can be used in diverse platforms like quadrotors, fix-wing airplanes, and ground vehicles with fast speed. 

In the future, running an experimental test with this BTO-RRT using a real quadrotor with GPS, cameras, IMU, and altimeter would be a good way to test the real-world performance of the proposed algorithm. And with the good characteristics of our proposed algorithm, it could greatly improve efficiency and enable better performance of tasks like SLAM, local navigation, and so on.

\addtolength{\textheight}{-12cm}   








\bibliography{references}{}

\begin{thebibliography}{10}

\bibitem{xu2022v2x}
Runsheng Xu, Hao Xiang, Zhengzhong Tu, Xin Xia, Ming-Hsuan Yang, and Jiaqi Ma.
\newblock V2x-vit: Vehicle-to-everything cooperative perception with vision
  transformer.
\newblock {\em arXiv preprint arXiv:2203.10638}, 2022.

\bibitem{xu2022opv2v}
Runsheng Xu, Hao Xiang, Xin Xia, Xu~Han, Jinlong Li, and Jiaqi Ma.
\newblock Opv2v: An open benchmark dataset and fusion pipeline for perception
  with vehicle-to-vehicle communication.
\newblock In {\em 2022 International Conference on Robotics and Automation
  (ICRA)}, pages 2583--2589. IEEE, 2022.

\bibitem{cao2022beaglerover}
Pengcheng Cao, James Strawson, Xuebin Zhu, Everbrook Zhou, Chase Lazar,
  Dominique Meyer, Zhaoliang Zheng, Thomas Bewley, and Falko Kuester.
\newblock Beaglerover: An open-source 3d-printable robotic platform for
  engineering education and research.
\newblock In {\em AIAA SCITECH 2022 Forum}, page 1914, 2022.

\bibitem{yang2016survey}
Liang Yang, Juntong Qi, Dalei Song, Jizhong Xiao, Jianda Han, and Yong Xia.
\newblock Survey of robot 3d path planning algorithms.
\newblock {\em Journal of Control Science and Engineering}, 2016, 2016.

\bibitem{Lavalle98rapidly-exploringrandom}
Steven~M. Lavalle.
\newblock Rapidly-exploring random trees: A new tool for path planning.
\newblock Technical report, 1998.

\bibitem{lavalle2001rapidly}
Steven~M LaValle and James~J Kuffner.
\newblock Rapidly-exploring random trees: Progress and prospects.
\newblock {\em Algorithmic and computational robotics: new directions},
  (5):293--308, 2001.

\bibitem{kuffner2000rrt}
James~J Kuffner and Steven~M LaValle.
\newblock Rrt-connect: An efficient approach to single-query path planning.
\newblock In {\em Proceedings 2000 ICRA. Millennium Conference. IEEE
  International Conference on Robotics and Automation. Symposia Proceedings
  (Cat. No. 00CH37065)}, volume~2, pages 995--1001. IEEE, 2000.

\bibitem{karaman2011sampling}
Sertac Karaman and Emilio Frazzoli.
\newblock Sampling-based algorithms for optimal motion planning.
\newblock {\em The international journal of robotics research}, 30(7):846--894,
  2011.

\bibitem{jordan2013optimal}
Matthew Jordan and Alejandro Perez.
\newblock Optimal bidirectional rapidly-exploring random trees.
\newblock 2013.

\bibitem{noreen2016comparison}
Iram Noreen, Amna Khan, and Zulfiqar Habib.
\newblock A comparison of rrt, rrt* and rrt*-smart path planning algorithms.
\newblock {\em International Journal of Computer Science and Network Security
  (IJCSNS)}, 16(10):20, 2016.

\bibitem{hess2016srrt}
Robin Hess, Roland Jerg, Tobias Lindeholz, Daniel Eck, and Klaus Schilling.
\newblock Srrt*-a probabilistic optimal trajectory planner for problematic area
  structures.
\newblock {\em IFAC-PapersOnLine}, 49(30):331--336, 2016.

\bibitem{kuwata2009real}
Yoshiaki Kuwata, Justin Teo, Gaston Fiore, Sertac Karaman, Emilio Frazzoli, and
  Jonathan~P How.
\newblock Real-time motion planning with applications to autonomous urban
  driving.
\newblock {\em IEEE Transactions on control systems technology},
  17(5):1105--1118, 2009.

\bibitem{dong2016rrt}
Yiqun Dong, Changhong Fu, and Erdal Kayacan.
\newblock Rrt-based 3d path planning for formation landing of quadrotor uavs.
\newblock In {\em 2016 14th International Conference on Control, Automation,
  Robotics and Vision (ICARCV)}, pages 1--6. IEEE, 2016.

\bibitem{zheng2020point}
Zhaoliang Zheng, T.R. BEWLEY, and Falko Kuester.
\newblock Point cloud-based target-oriented 3d path planning for uavs.
\newblock In {\em 2020 International Conference on Unmanned Aircraft Systems
  (ICUAS)}. IEEE, 2020.

\bibitem{bartels1995introduction}
Richard~H Bartels, John~C Beatty, and Brian~A Barsky.
\newblock {\em An introduction to splines for use in computer graphics and
  geometric modeling}.
\newblock Morgan Kaufmann, 1995.

\bibitem{sprunk2008planning}
Christoph Sprunk.
\newblock Planning motion trajectories for mobile robots using splines.
\newblock 2008.

\bibitem{rusu2007towards}
Radu~Bogdan Rusu, Nico Blodow, Zoltan Marton, Alina Soos, and Michael Beetz.
\newblock Towards 3d object maps for autonomous household robots.
\newblock In {\em 2007 IEEE/RSJ International Conference on Intelligent Robots
  and Systems}, pages 3191--3198. IEEE, 2007.

\bibitem{vasudevan2007cognitive}
Shrihari Vasudevan, Stefan G{\"a}chter, Viet Nguyen, and Roland Siegwart.
\newblock Cognitive maps for mobile robots—an object based approach.
\newblock {\em Robotics and Autonomous Systems}, 55(5):359--371, 2007.

\bibitem{zhang2021hierarchical}
Li~Zhang, Faezeh Tafazzoli, Gunther Krehl, Runsheng Xu, Timo Rehfeld, Manuel
  Schier, and Arunava Seal.
\newblock Hierarchical road topology learning for urban map-less driving.
\newblock {\em arXiv preprint arXiv:2104.00084}, 2021.

\bibitem{xu2021holistic}
Runsheng Xu, Faezeh Tafazzoli, Li~Zhang, Timo Rehfeld, Gunther Krehl, and
  Arunava Seal.
\newblock Holistic grid fusion based stop line estimation.
\newblock In {\em 2020 25th International Conference on Pattern Recognition
  (ICPR)}, pages 8400--8407. IEEE, 2021.

\bibitem{gao2016online}
Fei Gao and Shaojie Shen.
\newblock Online quadrotor trajectory generation and autonomous navigation on
  point clouds.
\newblock In {\em 2016 IEEE International Symposium on Safety, Security, and
  Rescue Robotics (SSRR)}, pages 139--146. IEEE, 2016.

\bibitem{fedorenko2018global}
Roman Fedorenko, Aidar Gabdullin, and Anna Fedorenko.
\newblock Global ugv path planning on point cloud maps created by uav.
\newblock In {\em 2018 3rd IEEE International Conference on Intelligent
  Transportation Engineering (ICITE)}, pages 253--258. IEEE, 2018.

\bibitem{bentley1975multidimensional}
Jon~Louis Bentley.
\newblock Multidimensional binary search trees used for associative searching.
\newblock {\em Communications of the ACM}, 18(9):509--517, 1975.

\bibitem{kakde2005range}
Hemant~M Kakde.
\newblock Range searching using kd tree.
\newblock {\em from the citeseerx database on the World Wide Web:
  http://citeseerx. ist. psu. edu/viewdoc/summary}, 2005.

\end{thebibliography}
\bibliographystyle{unsrt}

\end{document}